\RequirePackage{fix-cm}
\documentclass[twocolumn]{svjour3}          
\smartqed  
\usepackage{cite}
\usepackage{amsmath,amssymb,amsfonts}
\usepackage{graphicx}
\usepackage{textcomp}
\usepackage{subcaption}
\usepackage{algorithm,algpseudocode}
\usepackage{algorithmicx}
\usepackage{multirow}
\usepackage{enumitem}
\usepackage{bm}

\makeatletter
\DeclareRobustCommand{\textsupsub}[2]{{%
  \m@th\ensuremath{%
    ^{\mbox{\scriptsize\fontsize\sf@size\z@#1}}%
    _{\mbox{\scriptsize\fontsize\sf@size\z@#2}}%
  }%
}}
\makeatother
\journalname{Neural Computing and Applications}
\begin{document}

\title{SELM: Siamese Extreme Learning Machine with Application to Face Biometrics\thanks{Funding from King Mongkut's Institute of Technology Ladkrabang and project BIBECA (RTI2018-101248-B-I00 MINECO/FEDER)}
}

\titlerunning{SELM: Siamese Extreme Learning Machine with Application to Face Biometrics}        

\author{Wasu Kudisthalert\textsuperscript{1} \and Kitsuchart Pasupa\textsuperscript{1} \and Aythami Morales\textsuperscript{2} \and Julian Fierrez\textsuperscript{2}
}
\authorrunning{Wasu Kudisthalert et al.} 

\institute{Wasu Kudisthalert \\ \email{60606001@kmitl.ac.th} \\ \\
Kitsuchart Pasupa (Corresponding Author) \\ \email{kitsuchart@it.kmitl.ac.th} \\ \\
Aythami Morales \\ \email{aythami.morales@uam.es} \\ \\
Julian Fierrez \\ \email{julian.fierrez@uam.es} \\ \\
\textsuperscript{1} Faculty of Information Technology, King Mongkut's Institute of Technology Ladkrabang, Bangkok 10520, Thailand \\ \\
\textsuperscript{2} Biometric and Data Pattern Analytics Lab, Universidad Autonoma de Madrid, Spain
}

\date{Received: date / Accepted: date}

\maketitle

\begin{abstract}
Extreme Learning Machine is a powerful classification method very competitive existing classification methods. It is extremely fast at training. Nevertheless, it cannot perform face verification tasks properly because face verification tasks require comparison of facial images of two individuals at the same time and decide whether the two faces identify the same person. The structure of Extreme Leaning Machine was not designed to feed two input data streams simultaneously, thus, in 2-input scenarios Extreme Learning Machine methods are normally applied using concatenated inputs. However, this setup consumes two times more computational resources and it is not optimized for recognition tasks where learning a separable distance metric is critical. For these reasons, we propose and develop a Siamese Extreme Learning Machine (SELM). SELM was designed to be fed with two data streams in parallel simultaneously. It utilizes a dual-stream Siamese condition in the extra Siamese layer to transform the data before passing it along to the hidden layer. Moreover, we propose a Gender-Ethnicity-Dependent triplet feature exclusively trained on a variety of specific demographic groups. This feature enables learning and extracting of useful facial features of each group. Experiments were conducted to evaluate and compare the performances of SELM, Extreme Learning Machine, and DCNN. The experimental results showed that the proposed feature was able to perform correct classification at $97.87\%$ accuracy and $99.45\%$ AUC. They also showed that using SELM in conjunction with the proposed feature provided $98.31\%$ accuracy and $99.72\%$ AUC. They outperformed the well-known DCNN and Extreme Leaning Machine methods by a wide margin.
\end{abstract}

\section{Introduction}
\label{sec:introduction}

In the period of COVID-19 pandemic, A New Normal was introduced. People all around the world had to change their daily habits. They had to be constantly aware of their surroundings and had to keep everything around them clean of the virus all the time. The traveling history of every suspected COVID vector in an area had to be retraced when an infected person was detected in the area during that time, e.g., everyone arriving or leaving a building or community at the same time. To be able to retrace traveling history, accurate personal identification is of utmost importance. At this time of writing, some communities required visitors to identify themselves correctly before they were permitted an access into the communities. There are several ways to identify an individual, such as from their ID card, passport, fingerprint, iris or DNA~\cite{fierrez06phd,jain16years}, but one of the most convenient ways in many setups (like the discussed moving travellers due to COVID-19) is facial identification. At this time, numerous monitoring cameras have already been installed almost everywhere, such as in department stores, airports, border crossing facilities, cities and transportation stations, as a security and surveillance measure. An accurate and reliable face identification algorithm is required to identify individuals by their facial features \cite{patel18spm,fierrez21faceq}. The process of identification from facial features is a type of one-to-many mapping process, i.e., an unknown face is identified between multiple faces already registered in a database. The identification is assisted by taking into account demographic information---identity, age, gender, and ethnicity~\cite{tome15soft,sosa18cots,guo2020learning,ter21bias}. On the other hand, a face verification task is a one-to-one mapping process. The task verifies whether the individual with the recognized face is the same person registered in a system~\cite{sun2013hybrid}. This task is often used for authorizing a system, for example, for authorizing an access to a mobile device or a laptop \cite{patel20qid}. The advantage of this method over others like fingerprint recognition \cite{alonso09finger} is that it does not require anyone touching anything \cite{fierrez18touch}.

Face recognition techniques have been developed for decades~\cite{galbally2019study}, e.g., Geometric based approaches~\cite{shi2006effective}, Local feature analysis~\cite{arca2003face}, Dictionary based learning~\cite{chen2012dictionary,patel2012dictionary}, Hand-crafted features~\cite{jin2014hand,antipov2015learned} and, recently, Deep Convolutional Neural Network (DCNN)~\cite{yuan2017convolutional}. Recently, many large-scale face datasets containing millions of images have been available~\cite{8599059,kemelmacher2016megaface,cao2018vggface2} for training deep learning model. Nevertheless, the class distributions of some variates in those datasets were rather imbalanced, causing statistical bias \cite{ter21bias}. This issue was associated with imbalanced representation of classes in a dataset. An effect of the bias was reported in~\cite{phillips2011other}. They reported that algorithms invented by Asian researchers were able to distinguish Asian subjects better than Caucasian subjects. Conversely, algorithms from the West performed better on Caucasian subjects. Along the same line, a study by~\cite{buolamwini2018gender} reported that a commercial face recognition system yielded better outcomes on male individuals and lighter individuals, but worse outcomes on darker females. Therefore, bias in class proportion and demographic variates would strongly affect a biometric system performance \cite{serna21insidebias}. This concern could be alleviated by utilizing datasets evenly distributed across demographics~\cite{klare2012face,serna2020sensitiveloss}. Training a model on a specific group could reduce data diversity and allows the model to learn better characteristics of each class. Interestingly, the performance of a model that was intensely trained on a very specific group, like male and female or every different ethnic in an area, might be superior than the performance of a conventionally trained model \cite{acien18bias}.

Face representation is an important part of the face verification task. Historically, different representation techniques have been used to extract facial information from face images. In the past, computer vision hand-crafted techniques were employed to transform face images into useful features such as geometry-based features that utilized the shape of a face and its landmarks to represent the appearance of the face and its components. At the time of writing, the most competitive face representations are obtained using Deep Convolutional Neural Networks (DCNN) optimized according to different loss functions~\cite{TripletLoss,SphereFace,arcFace}. Among the different loss functions, triplet loss (a triplet network) is a Distance-Metric approach designed as a type of Siamese network~\cite{hoffer2015deep}. This triplet network has a hierarchy that starts learning from low level features to high level features, i.e., from pixels to classes. It could be fed with two inputs in parallel. A pair of faces can be fed into a triplet network to output a similarity/distance coefficient between the two input face images. The value of this coefficient is then usually compared against a threshold. An identity match is positive when it exceeds the threshold. Else, it was a mismatch. Fortunately, several machine learning algorithms could be employed to enhance the performance of the face verification task. They could learn the pattern of the data and distinguish them into classes instead of measuring the similarity/distance coefficient between two faces. Nevertheless, most of them could not deal with this task without some modification because their architecture was designed to be fed with one input at a time. Fortunately, this can be solved by linking two inputs into a concatenated input, but certain unavoidable bias would be introduced, e.g., the exact order of concatenation of the two inputs might introduce a bias--- a different order yields a different output. In this work, we restructured a well-known classification algorithm, Extreme Learning Machine (ELM)~\cite{huang2004extreme}, to accept twin inputs simultaneously and eliminate this kind of bias. The restructured algorithm was based on a single hidden-layer feedforward neural network (SLFN).

The following are the main contributions of the present paper:
\begin{itemize}
    \item We propose a novel classification method for verification tasks named Siamese Extreme Learning Machine (SELM). The proposed method adapts standard Extreme Learning Machine architectures in order to process parallel inputs in an efficient way.
    \item We develop a demographic-dependent triplet model that is shown to improve the performance in face verification.
    \item The proposed framework is demonstrated to distinguish gender, ethnicity, and face accurately.
    \item We carry out a performance comparison in face biometrics between biased and unbiased triplet models under different setups: subject-independent, gender-dependent, and gender-ethnicity-dependent.
    \item We carry out a performance comparison between Siamese and Non-Siamese algorithms.
\end{itemize}

\section{Related works}
\label{sec:related_works}

Some of the key challenges in face recognition are the following: 1) inadequate quality of facial images deteriorates the performance of face detection and verification \cite{fierrez21faceq}; and 2) biases between cohorts of people, specially with respect to privileged ones, deteriorates the performance of face recognition in general and introduces undesired discrimination between population groups \cite{sixta2020fairface,serna2019algorithmic}. There are many powerful and well-known techniques for face recognition \cite{patel18spm}. In this section on related works, we will first discuss the strengths and weaknesses of key techniques for face recognition with emphasis on the two challenges indicated above. We will then position our proposed machine learning methods in context.

\subsection{Demographic variates in face recognition}

Gender and race are two important demographic variates representing subject-specific characteristics of the human face. Other variates have also been proven useful for face recognition. For example, the skin tone can help improve face recognition performance. Back to demographic variates, Cook~\textit{et al.}~\cite{cook2019demographic} examined the effects of demographic variates on face recognition through leading commercial face biometric systems. They investigated the effects with a dataset of 363 subjects in a controlled environment and found that many demographic covariates significantly affected the face recognition performance, including gender, age, eyewear, height, and especially skin reflectance. Lower skin reflectance (darker skin tone) was associated with lower efficiency (longer transaction time) and accuracy, in terms of mated similarity score. The study also revealed that skin reflectance was a significantly better predictor than self-identified race variates. Buolamwini and Gebru~\cite{buolamwini2018gender} reported a significant bias in well-known commercial gender classification systems, i.e., Microsoft~\cite{del2018introducing}, IBM~\cite{high2012era}, and Face++. They found that darker-skinned females were the most misclassified group with an error rate of $34.7\%$, while the mis-classified rate of lighter-skinned males was only $0.8\%$. The largest difference in error rate between the best and the worst classified groups was $34.4\%$. They concluded that these three classification systems yielded the best accuracy for lighter-skinned individuals and males but the worst accuracy for darker-skin females due to the mentioned bias. Several studies have reported that Caucasian and male individuals are easier to distinguish by face recognition algorithm~\cite{klare2012face,buolamwini2018gender,cook2019demographic}. Recently, Lu~\textit{et al.}~\cite{8599059} investigated the effects of demographic groups on face recognition and found that the difficulty of unconstrained face verification varies significantly with different demographic variates. Males are easier to verify than females, and old subjects are recognized better than young individuals. On the other hand, light-pink skin tone is recognized with the best performance. Moreover, gender and skin tone variates are not significantly correlated.

On the other hand, some works have exploited the inherent differences between population groups for stronger and more fair recognition. Phillips \textit{et al.} \cite{phillips2011other} and O'Toole \textit{et al.} \cite{OTOOLE2012169} showed the importance of demographic composition and modeling. They reported that recognition of face identities from a homogeneous population (same-race distribution) was easier than recognition from a heterogeneous population. Liu~\textit{et al.}~\cite{lui2009meta} showed that the recognition performance using a training set that contained facial images of Caucasians and East Asians at a ratio of 3:1 was better at identifying East Asians in every case. Klare~\textit{et al.}~\cite{klare2012face} and Vera-Rodriguez~\textit{et al.}~\cite{Vera-Rodriguez_2019_CVPR_Workshops} improved face-matching accuracy by training exclusively on specific demographic cohorts of which demographic variates were evenly distributed. This solution could reduce face bias and offer higher accuracy across all demographic cohorts. Vera-Rodriguez~\textit{et al.}~\cite{Vera-Rodriguez_2019_CVPR_Workshops} proposed a gender-dependent training approach to improve face verification performance that reduced the effect of gender as a recognition covariate. The approach improved AUC performance from 94.0 to 95.2. Vera-Rodriguez~\textit{et al.}~\cite{Vera-Rodriguez_2019_CVPR_Workshops} and Serna~\textit{et al.}~\cite{serna2019algorithmic, serna2020sensitiveloss} applied deap learning methods to train face recognition models and benchmarked the models over multiple privileged classes. Conventional methods (not exploiting data diversity) resulted in poor performance when demographic diversity was large. Their experimental results showed a big performance gap between the best class (Male-White) and the worst class (Female-Black) that reached up to $200\%$. The above studies also demonstrated that training the models on specific demographic cohorts can be a possible solution to those large performance differences between cohorts. For example, useful features for distinguishing black individuals may be different to those for white individuals. Thus, training a model with specific groups of individuals may direct the model to better learn special characteristics of the groups.

Many well-known large-scale face recognition datasets have been published, such as MS-Celeb-1M \cite{8599059}, Megaface \cite{kemelmacher2016megaface}, and VGGFace2 \cite{cao2018vggface2}. These datasets contain more than a million face images, but most of them are highly-biased datasets, composed mainly of Caucasian people ($70\%+$), while $40\%+$ come from a Male-Caucasian cohort. Recently, Wang \textit{et al.} \cite{wang2019racial,wang2020mitigating} introduced diverse and discrimination-aware face databases with even-distributed populations: Asian, Black, Caucasian, and Indian. However, they did not balance the gender distribution. Along the same line, Morales~\textit{et al.} \cite{morales21sensitivenets} introduced the DiveFace database with equal distribution for six demographic groups: Female-Asian, Male-Asian, Female-Black, Male-Black, Female-Caucasian, and Male-Caucasian. The dataset was designed to be unbiased in terms of Gender and Ethnicity, which is useful both for training fair recognizers and evaluating them in terms of fairness across population groups.

\subsection{Machine learning architectures for face recognition}

\begin{figure}[t]
    \centering
	\includegraphics[width=0.89\columnwidth]{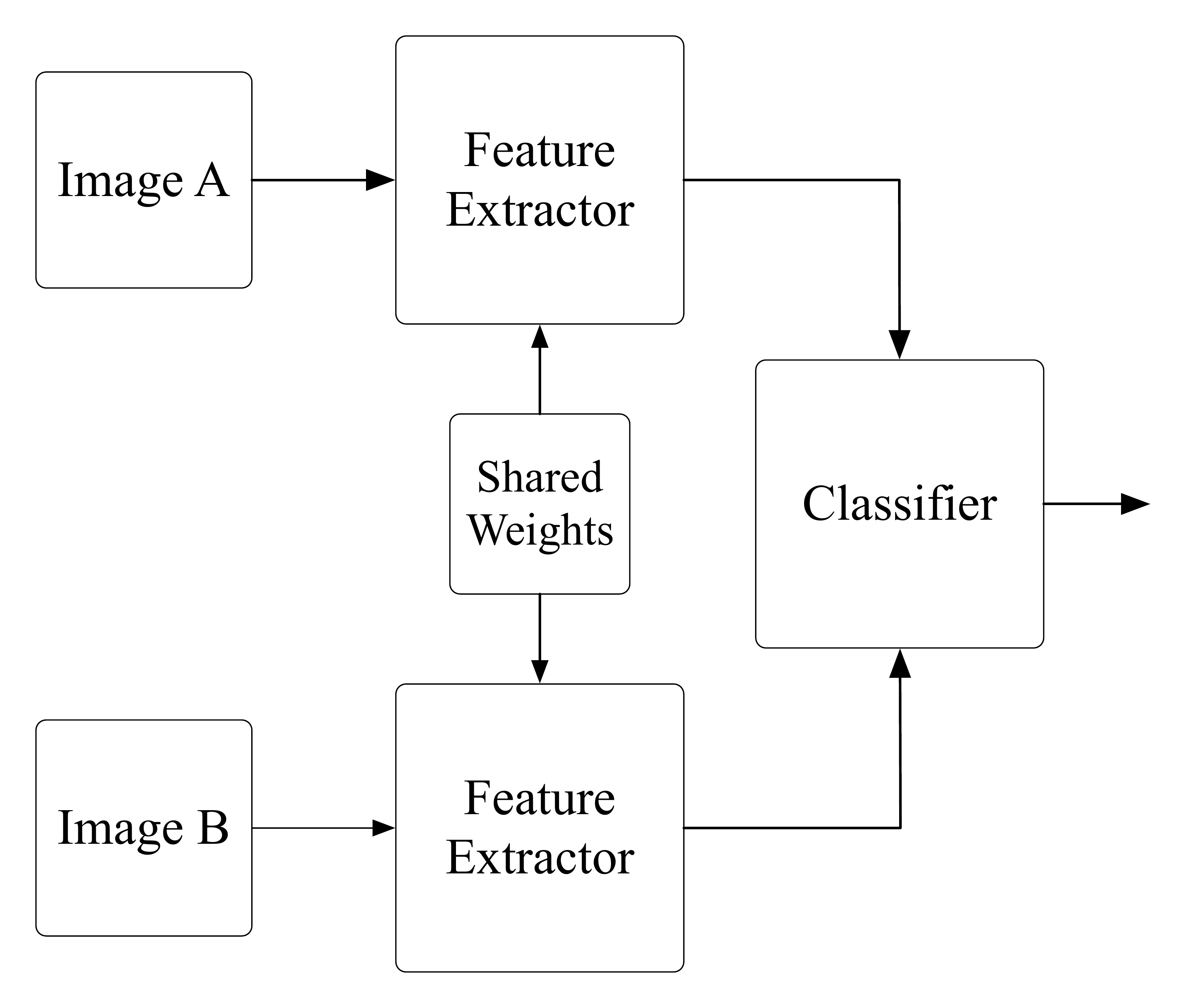}
	\caption{Siamese network concept.}
	\label{fig:siamese_concept}
\end{figure}

Machine learning classification techniques have been popular for face recognition tasks. Successful algorithms are for example: Support Vector Machines (SVM) \cite{dadi2016improved}, Extreme Learning Machines (ELM) \cite{gurpinar2016kernel}, Random Forests \cite{liu2018conditional}, and Deep Convolutional Neural Networks~\cite{wen2016discriminative}, the last one now dominating the field. Goswami~\textit{et al.}~\cite{goswami2017face} summarized the performances of features extracted by deep and shallow feature extractor approaches. The experimental results showed clearly the superiority of deep features. Other works such as Liu~\textit{et al.}~\cite{liu2018conditional}, Bianco~\cite{bianco2017large} and Wong~\textit{et al.}~\cite{wong2019realization} have also shown the robustness and improved recognition of face biometrics based on features extracted from DCNNs. However, the typical classification architecture in those works was designed to be fed with one input image at a time. For comparing two input faces (e.g., for authentication) there is a need for extending the basic DCNN architecture to process two inputs.

One popular approach to exploit a DCNN backbone for comparing two inputs is the Siamese architecture. The concept is to train a feature representation by comparing pairs of facial images. The conceptual diagram is shown in Figure~\ref{fig:siamese_concept}. In the present paper we will adopt this architecture in combination with an Extreme Learning Machine (cf. Section~\ref{sec:elm} for an introduction to this type of networks.) ELMs have been shown to be quite successful in various tasks related to face biometrics, but so far Siamese architectures have not been explored yet for enhancing basic ELM methods.

As example of ELMs for face biometrics, Laiadi~\textit{et al.}~\cite{laiadi2019kinship} predicted kinship relationship by comparing facial appearances. They used three different types of features: deep features using VGG-Face model, BSIF-Tensor, and LPQ-Tensor features using MSIDA. These three features of the two considered face images were then measured by cosine similarity, then the measured data were concatenated as a vector for computing a kinship score by ELM. The proposed approach was up to $3\%$ more accurate than a baseline ResNet-based method. Wong~\textit{et al.}~\cite{wong2019realization} adopted ELM to tackle face verification. They added a top layer of DeepID~\cite{sun2014deep} with ELM as the classification layer instead of a soft-max layer. This approach improved the accuracy by $1.32\%$ and $26.33\%$, respectively, for a conventional DeepID and ELM.

In the present paper, we develop and explore a novel Siamese classification algorithm for face verification with ELM backbone. The proposed algorithm compares pairs of facial images based on demographic traits. The traits are used as factors for selecting feature extraction models. The main aim of this work is to boost the performance of the algorithm by decreasing the verification errors. A secondary aim of this work is to investigate the dependency of the performance on demographic variates.

\section{Methods}
\label{sec:methods}

\subsection{Extreme Learning Machine}
\label{sec:elm}
Extreme Learning Machines (ELMs) were first introduced by Huang~\textit{et al.}~\cite{huang2004extreme}. They are based on a single hidden-layer feedforward neural network (SLFN) architecture of which the weights are obtained by the closed-form solution of an inverse problem, instead of the typical iterative back-propagation optimization. It has been demonstrated that this closed-form solution in ELMs yields a small classification error and extremely fast learning. The ELM architecture consists of $m$ input neurons ($m=$ input dimensions). The input neurons are fully connected with $l$ hidden neurons each one with weighted inputs according to $\bm{\mathrm{w}}_{i}$, with $i=1,\ldots,l$. The weights between the hidden layer and the output layer are defined as hidden layer output weights $\bm{\mathrm{\beta}}$ that are used to determine the prediction outputs $\bm{\mathrm{\hat{y}}}$. The model is expressed mathematically as (scalars in italics, column vectors in bold lowercase, matrices in bold uppercase, $^\intercal$ denotes transpose):

\begin{equation}
\hat{y}_{j}=\sum_{i=1}^{l} \bm{\mathrm{\beta}}_{i}g(\bm{\mathrm{w}}_{i}^{\intercal} \cdot \bm{\mathrm{x}}^{j}+b_{i}),\textrm{ for }j=1,\ldots,n.
\end{equation}

The output from the hidden layer is processed by an activation function $g(\cdot)$ with a linear combination of input $\bm{\mathrm{X}}$ and synaptic weights $\bm{\mathrm{w}}$ as well as bias $b$, where $\bm{\mathrm{X}} \in \mathcal{R}^{n\times m}$ and $n$ is the number of input samples. It should be noted that the set of $\bm{\mathrm{w}}$ and $b$ are randomly generated once to speed up the training process. Therefore, the activity of the hidden node can be written as:

\begin{align}
\label{matrixH}
{\bm{\mathrm{H}}}=
\begin{bmatrix}
\bm{\mathrm{h}}^{\intercal}(\bm{\mathrm{x}}^{1})\\
\vdots \\
\bm{\mathrm{h}}^{\intercal}(\bm{\mathrm{x}}^{n})
\end{bmatrix}=
\begin{bmatrix}
g(\bm{\mathrm{w}}_{1}^{\intercal} \bm{\mathrm{x}}^{1} + b_{1}) & \dots & g(\bm{\mathrm{w}}_{l}^{\intercal} \bm{\mathrm{x}}^{1} + b_{l}) \\
\vdots & \ddots & \vdots \\
g(\bm{\mathrm{w}}_{1}^{\intercal} \bm{\mathrm{x}}^{n} + b_{1}) & \dots & g(\bm{\mathrm{w}}_{l}^{\intercal} \bm{\mathrm{x}}^{n} + b_{l})
\end{bmatrix}_{\, n \, \times \, l}.
\end{align}

The prediction score is then expressed by:

\begin{equation}
\hat{\bm{\mathrm{y}}}=\bm{\mathrm{H}}\bm{\mathrm{\beta}}.
\end{equation}

ELM minimizes the mean square error between true target labels $\bm{\mathrm{y}}$ and predicted targets $\bm{\mathrm{\hat{y}}}$ by using the following objective function:

\begin{equation}
\min_{\bm{\mathrm{\beta}}} \frac{1}{2} \parallel \bm{\mathrm{\hat{y}}} - \bm{\mathrm{y}} \parallel ^{2} _{2}.
\end{equation}

The optimal solution of the hidden layer output weights $\bm{\mathrm{\beta}}$ is finally calculated by the Moore-Penrose pseudo-inverse:

\begin{equation}
\bm{\mathrm{\beta}} = ({\bm{\mathrm{H}}}^{\intercal} {\bm{\mathrm{H}}})^{-1}{\bm{\mathrm{H}}}^{\intercal}{\bm{\mathrm{y}}}.
\end{equation}

\subsection{Weighted Similarity Extreme Learning Machine}
\label{sec:welm}

\begin{figure}[t]
    \centering
	\includegraphics[width=\columnwidth]{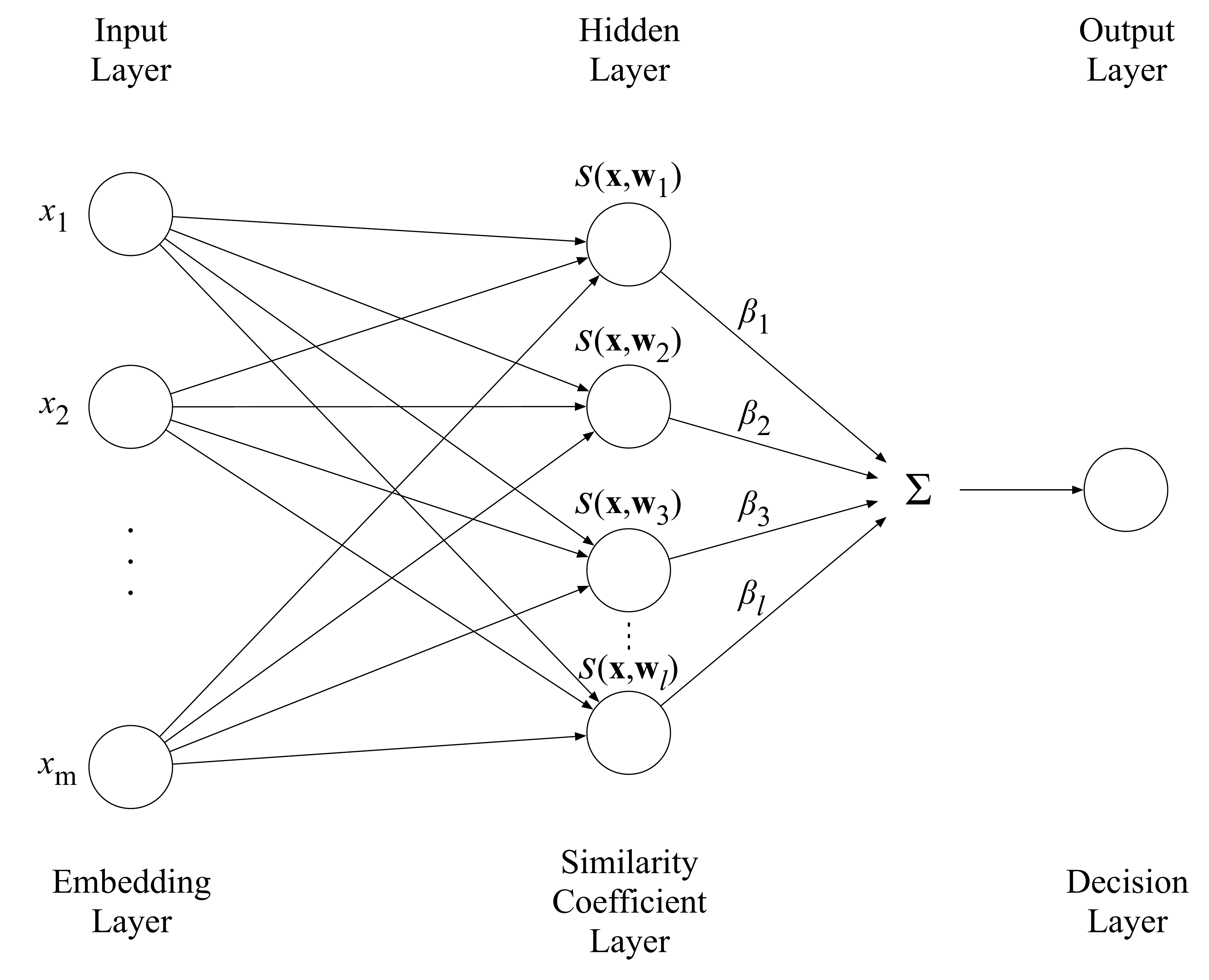}
	\caption{Weighted Similarity Extreme Learning Machine architecture.}
	\label{fig:welm_architecture}
\end{figure}

The WELM architecture is shown in Figure~\ref{fig:welm_architecture}, where the conventional activation function $g(\cdot)$, e.g., sigmoid or radial basis function, is replaced with a similarity-based activation function $s(\cdot)$, e.g., cosine similarity or Euclidean distance. WELM can reduce training time because it does not need any tuning of the kernel parameters. It yields better performance especially when dealing with similarity-based tasks~\cite{pasupa2018virtual,kudisthalert2020counting}. In WELM, the $\bm{\mathrm{H}}$ matrix in conventional activation is replaced by:

\begin{align}
\label{matrixSH}
{\bm{\mathrm{H}}}=
\begin{bmatrix}
\bm{\mathrm{h}}^{\intercal}(\bm{\mathrm{x}}^{1})\\
\vdots \\
\bm{\mathrm{h}}^{\intercal}(\bm{\mathrm{x}}^{n})
\end{bmatrix}=
\begin{bmatrix}
s(\bm{\mathrm{x}}^{1},\bm{\mathrm{w}}_{1}) & \dots & s(\bm{\mathrm{x}}^{1},\bm{\mathrm{w}}_{l}) \\
\vdots & \ddots & \vdots \\
s(\bm{\mathrm{x}}^{n},\bm{\mathrm{w}}_{1}) & \dots & s(\bm{\mathrm{x}}^{n},\bm{\mathrm{w}}_{l})
\end{bmatrix}_{\, n \, \times \, l}.
\end{align}

The weights $\bm{\mathrm{w}}$ are randomly selected from a training set $\bm{\mathrm{X}}$, thus, $\bm{\mathrm{w}} \subset \bm{\mathrm{X}}$.

\subsection{Siamese Extreme Learning Machine}
\label{sec:selm}

\begin{figure*}[t]
    \centering
	\includegraphics[width=0.8\textwidth]{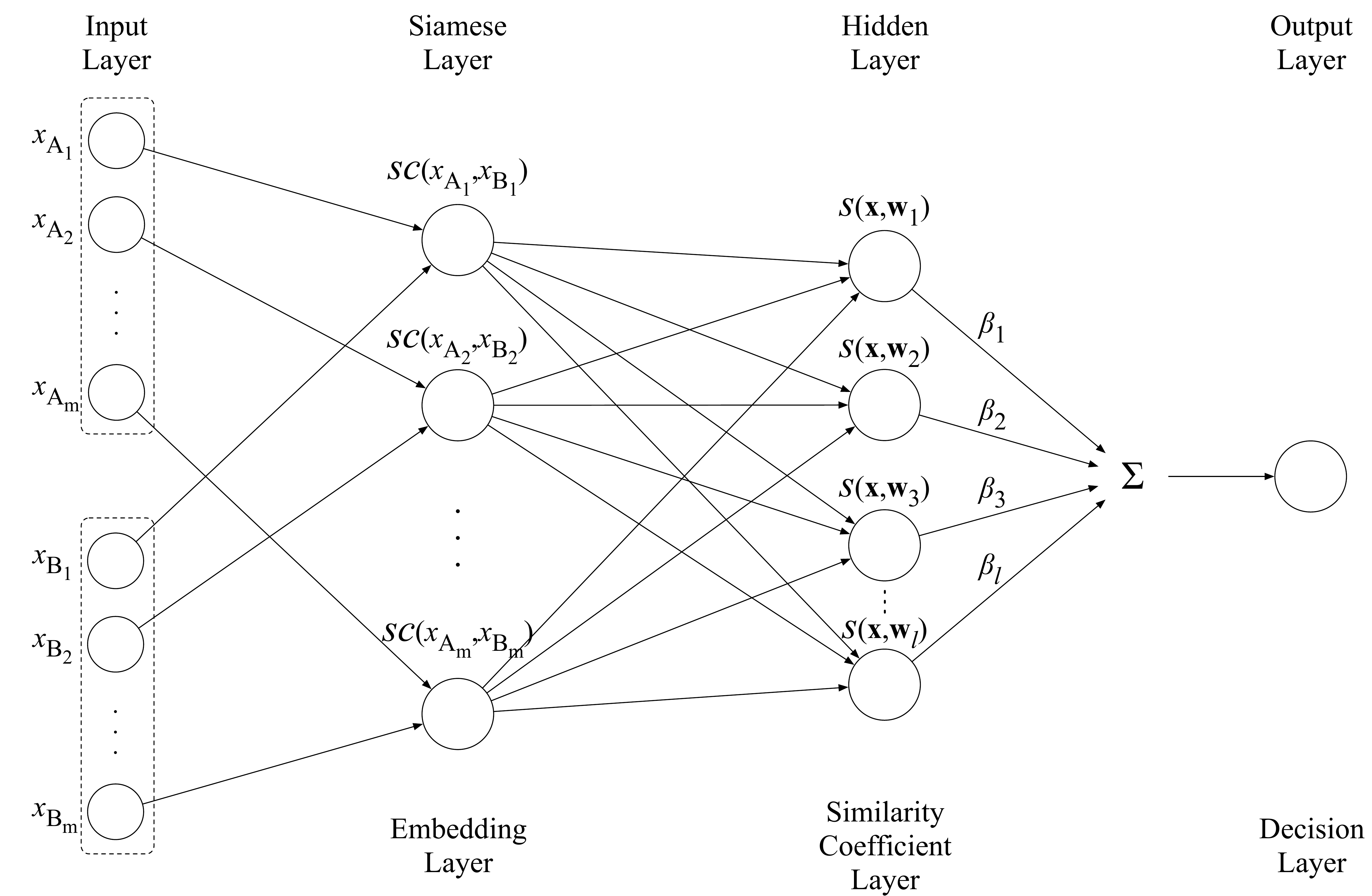}
	\caption{Siamese Extreme Learning Machine architecture.}
	\label{fig:selm_architecture}
\end{figure*}

This paper proposes a novel Siamese Extreme Learning Machine (SELM) architecture to handle verification tasks that require simultaneous comparison of two identities. SELM is developed on a WELM network backbone. Input vectors $\bm{\mathrm{x}}^{\text{A}}$ and $\bm{\mathrm{x}}^{\text{B}}$ from identity A and B, respectively, are fed into WELM after a Siamese input layer, turning the conventional WELM architecture into a SELM architecture capable of feeding two inputs simultaneously and in parallel into the network, as shown in Figure~\ref{fig:selm_architecture}.

A Siamese condition function $sc(\cdot)$ in the Siamese layer is the core of SELM. The function combines two input vectors using one of the following equations: 

\begin{itemize}
\item Summation condition function:
\begin{align}
\bm{\mathrm{x}}=\bm{\mathrm{x}}_{\text{A}}+\bm{\mathrm{x}}_{\text{B}},
\end{align}
\item Distance condition function:
\begin{align}
\bm{\mathrm{x}}=|\bm{\mathrm{x}}_{\text{A}}-\bm{\mathrm{x}}_{\text{B}}|,
\end{align}
\item Multiply (Hadamard product) condition function:
\begin{align}
\bm{\mathrm{x}}=\bm{\mathrm{x}}_{\text{A}}\odot\bm{\mathrm{x}}_{\text{B}},
\end{align}
\item Mean condition function:
\begin{align}
\bm{\mathrm{x}}=\frac{\bm{\mathrm{x}}_{\text{A}}+\bm{\mathrm{x}}_{\text{B}}}{2}.
\end{align}
\end{itemize}

Note that this Siamese layer can be also interpreted as an initial feature-level information fusion stage\cite{fierrez18fusion}.

The pseudocodes of the training and prediction processes of SELM are shown in Algorithm~\ref{alg:selm_algorithm}.


\begin{algorithm}[t]
\caption{Siamese Extreme Learning Machine}
\label{alg:selm_algorithm}
\begin{algorithmic}[1]
\Function{SELM\_Train}{${\bm{\mathrm{X}}}^{\text{Train}}$, $\bm{\mathrm{y}}$}
\State{$n \gets$ \#samples in ${\bm{\mathrm{X}}}^{\text{Train}}$}
\State{$n_\text{Pairs} \gets$ total \#pairs of samples chosen from the $n$ available samples}
\State{$n_\text{P} \gets$ \#pairs in Positive class out of $n_\text{Pairs}$}
\State{$n_\text{N} \gets$ \#pairs in Negative class out of $n_\text{Pairs}$}
\State{${\bm{\mathrm{X}}}^{\text{Train,EL}} \gets n_\text{Pairs}$ selected from ${\bm{\mathrm{X}}}^{\text{Train}}$ after passing them across the Embedding Layer (see Fig.~\ref{fig:selm_architecture})}
\State{$\bm{\mathrm{W}}$ $\gets$ randomly select subset of $l$ rows of ${\bm{\mathrm{X}}}^{\text{Train,EL}}$ (which has $n_\text{Pairs}$ rows in total)}
\For{$ i \gets 1$ to $n_\text{Pairs}$}\Comment{balance imbalance dataset}
\State{$\gamma_{i}\gets\sqrt{\frac{\max(n_\text{P}, n_\text{N})}{\#\text{samples in }{y}_{i}\text{ class}}}$}
\EndFor
\State{$\hat{{\bm{\mathrm{H}}}} \gets$ Eq.~\ref{matrixSH} considering ${\bm{\mathrm{X}}}^{\text{Train,EL}}$, $\bm{\mathrm{W}}$, and each $s(\bm{\mathrm{x}}^i,\cdot)$ substituted for $\gamma_{i} \cdot s(\bm{\mathrm{x}}^i,\cdot)$}
\State{$\bm{\mathrm{\beta}} \gets (\frac{1}{C}+\hat{{\bm{\mathrm{H}}}}^{\intercal}\hat{{\bm{\mathrm{H}}}})^{-1}(\hat{{\bm{\mathrm{H}}}}^{\intercal}\bm{\mathrm{\gamma}} \cdot {\bm{\mathrm{y}}})$}
\State \Return{$\bm{\mathrm{W}}$, $\bm{\mathrm{\beta}}$}
\EndFunction
\item[]
\Function{SELM\_Predict}{$\bm{\mathrm{W}}$, $\bm{\mathrm{\beta}}$, ${\bm{\mathrm{x}}}^{\text{Test}}$}
\State{${\bm{\mathrm{h}}}^{\intercal} \gets \bm{\mathrm{h}}^{\intercal}({\bm{\mathrm{x}}}^{\text{Test}})$ like in Eq.~\ref{matrixSH} making use of $\bm{\mathrm{W}}$}
\State{$\hat{y} \gets {\bm{\mathrm{h}}}^{\intercal}\bm{\mathrm{\beta}}$}
\State \Return{$\hat{y}$} 
\EndFunction
\end{algorithmic}
\end{algorithm}

\subsection{Triplet Convolutional Neural Networks}
\label{sec:triplet_network}

\begin{figure}[t]
    \centering
	\includegraphics[width=0.85\columnwidth]{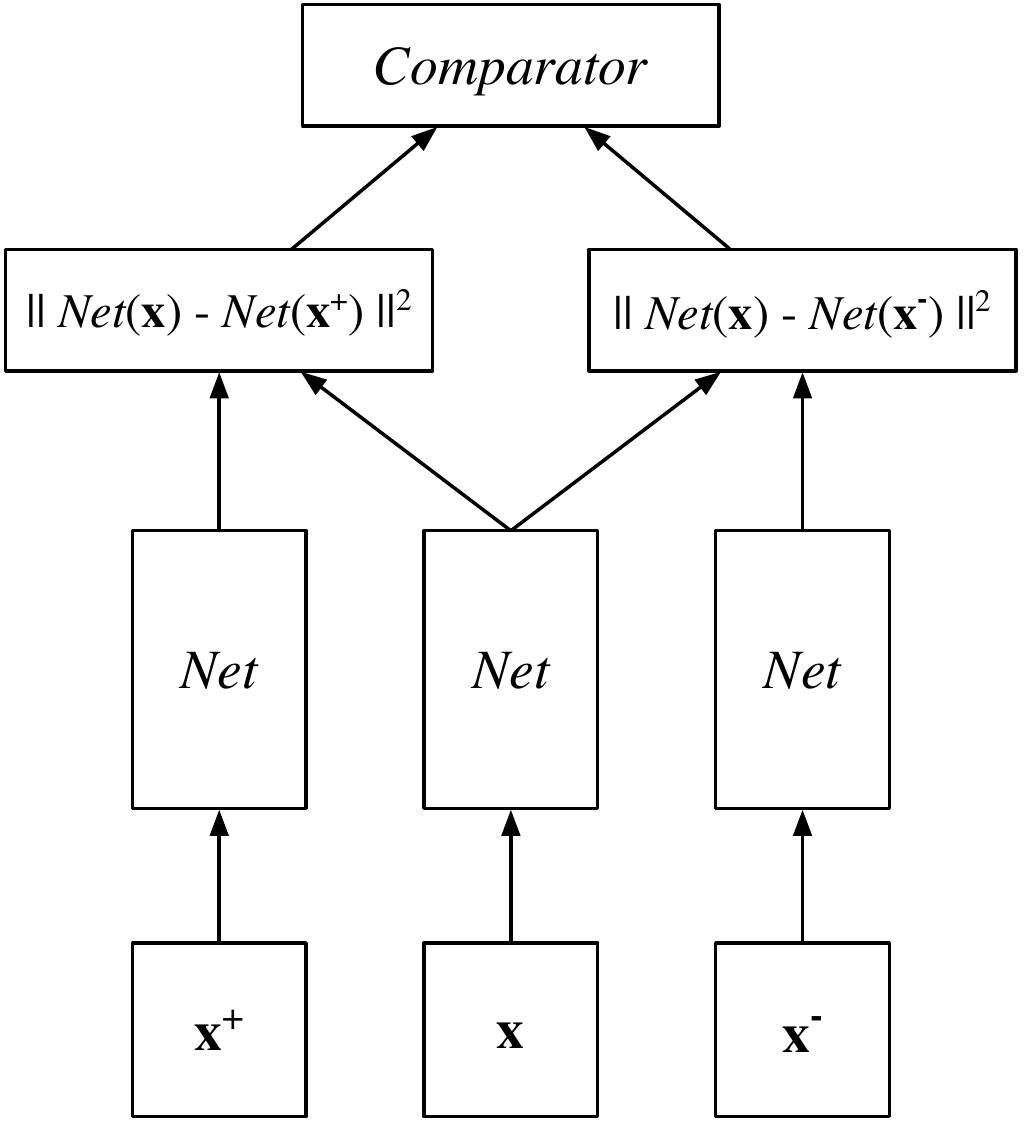}
	\caption{Triplet network structure (\emph{Net} can be a DCNN).}
	\label{fig:triplet_network_structure}
\end{figure}

The triplet network model was proposed for learning useful representations by distance comparisons~\cite{hoffer2015deep} between three samples: $anchor$ sample $\bm{\mathrm{x}}$, $positive$ sample $\bm{\mathrm{x}}^{+}$, and $negative$ sample $\bm{\mathrm{x}}^{-}$.The triplet network structure is shown in Figure~\ref{fig:triplet_network_structure}. As can be seen, the network employs DCNNs as backbone to optimize the weights of the model with back-propagation. These core networks are identical sharing the same weights. The aim of the triplet network is to minimize the $d_{p}$ distance between the $anchor$ and the $positive$ sample and to maximize the $d_{n}$ distance between the $anchor$ and the $negative$ sample. The $positive$ sample and the $anchor$ sample come from the same identity, while the $negative$ sample comes from a different identity. The Euclidean distance of $d_{p}$ and $d_{n}$ is expressed as,

\begin{equation}
d_{p}=\|Net(\bm{\mathrm{x}})-Net(\bm{\mathrm{x}}^{+})\|^{2}
\end{equation}
\begin{equation}
d_{n}=\|Net(\bm{\mathrm{x}})-Net(\bm{\mathrm{x}}^{-})\|^{2}.
\end{equation}

The triplet loss is then calculated as a loss function of the network as follows:

\begin{equation}
\begin{aligned}
\mathcal{L}_{triplet}=[d_{p}-d_{n}+\alpha]_{+},
\end{aligned}
\end{equation}

where the $\alpha$ parameter is a soft margin. The objective of the learning function is to satisfy $d_{n} \ge d_{p}+\alpha$. In this study, we trained a number of triplet networks with several demographic groups so that they could learn population-specific facial information.

\section{Proposed framework}
\label{sec:proposed_framework}

\begin{figure*}[t]
    \centering
	\includegraphics[width=\textwidth]{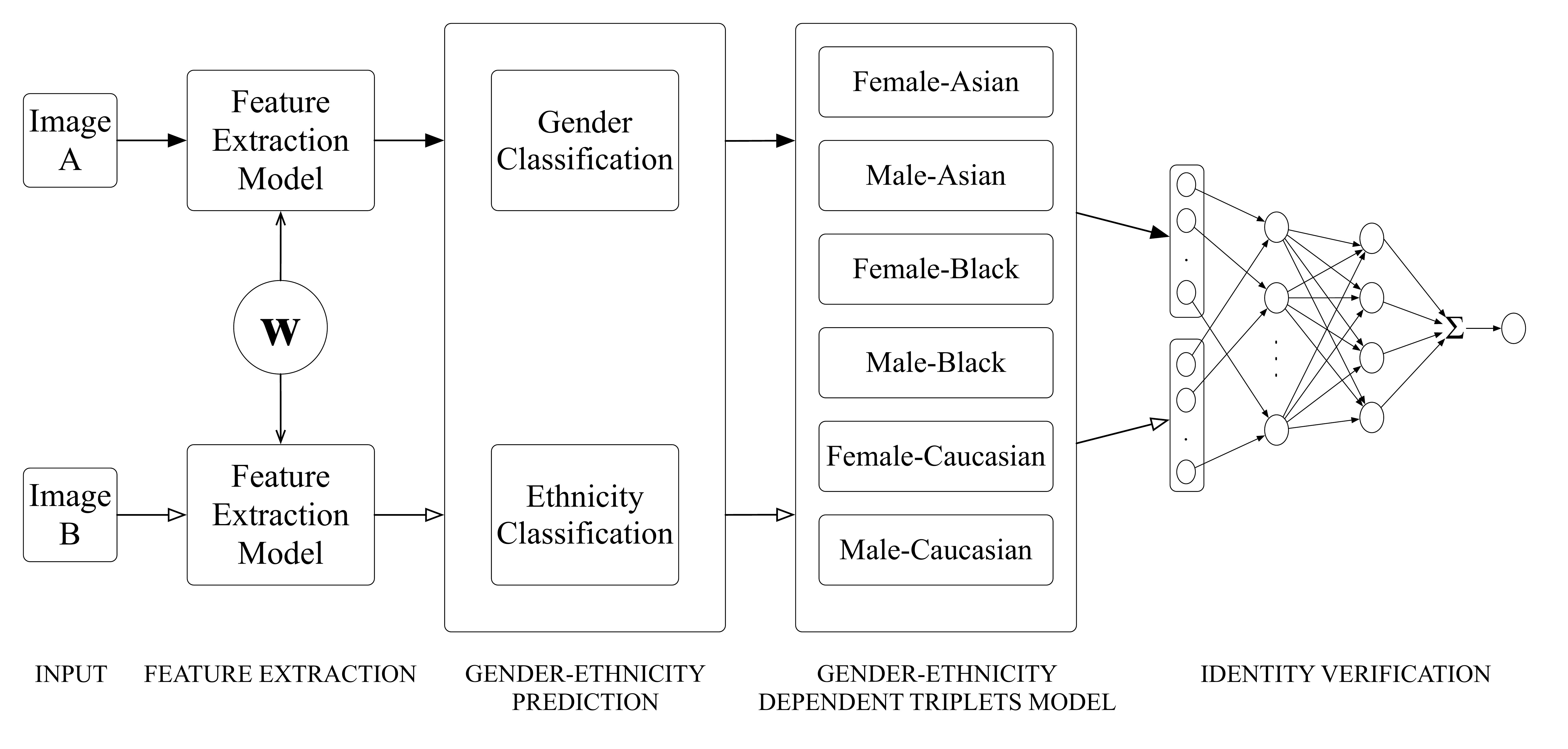}
	\caption{The workflow of the proposed framework.}
	\label{fig:framework}
\end{figure*}

The proposed framework is shown in Figure~\ref{fig:framework}. It consists of five stages. The framework was designed to verify the identity of two input facial images. The input images are first classified into gender and ethnicity to select a gender- and ethnicity-dependent triplet model for each input. The details of each stage are explained below.

\begin{enumerate}

    \item First stage (Input): input color facial images were first cropped and aligned properly~\cite{Deng_2020_CVPR} before being fed into the next stage. It should be noted that the two images passed in parallel through every process in the framework simultaneously. The input direction is shown by black or white line with arrow for Image A and B, respectively.
    
    \item Second stage (Feature Extraction): \mbox{ResNet-50} is a 50-layer-deep CNN with skip connections. It is one of the most robust methods for face recognition among existing deep architectures, such as \mbox{VGG-16}, \mbox{Inception-3} and \mbox{DenseNet-121}~\cite{vera2019facegenderid,serna2019algorithmic,wang2019benchmarking}. \mbox{ResNet-50} was used as the features extraction model. It was trained with a large-scale face dataset, VGGFace2~\cite{cao2018vggface2}. \mbox{ResNet-50} required a color image size of 224$\times$224 pixels as input. The length of the output was 2,048 features.
    
    \item Third stage (Gender-Ethnicity Prediction). This stage consists of two classification tasks: gender and ethnicity classification.
    
    \item Fourth stage (Gender-Ethnicity-Dependent Triplet Model): the extracted facial features from the second stage are used by one of six models to extract the triplets. Each triplet model was specially trained only with data in its Gender-Ethnicity-dependent class because, for example, a Female-Black person may have distinctive features different from those in the other classes. Thus, letting the model learn only in a specific class would make it better in recognizing the distinctive characteristics of the data in that class. In this work, we used the DiveFace dataset for training the triplet models because it is a discrimination-aware face dataset that provides the same distribution from the six different demographic groups considered here. Details of DiveFace are described in Section~\ref{sec:diveface}.
    
    \item Fifth stage (Identity Verification): there are two steps in this verification task. First, the pair of images A and B is classified as an impostor match if both images result in different Gender-Ethnicity classes in the third stage. Second, machine learning models are applied to verify if both images come from the same identity. In this work, we compare the proposed approach SELM to the performance of standard ELM and ResNet (which is now one of the most common DCNNs used for face recognition~\cite{laiadi2019kinship}). Incidentally, ResNet is also a core component of our proposed approach for training the triplet models.
    
\end{enumerate}

\section{Experimental protocol}
\label{sec:experimentalt_protocol}

\subsection{Dataset}
\label{sec:dataset}
In this study, we used two datasets: DiveFace and Labeled Faces in the Wild. DiveFace is a diversity-aware face recognition dataset for training models such as Gender classification, Ethnicity classification, and Gender-Ethnicity-dependent triplet models. Labeled Faces in the Wild dataset is a well-known large scale face dataset in the face recognition domain for performance evaluation.

\subsubsection{DiveFace: a diversity-aware face recognition dataset}
\label{sec:diveface}

\begin{figure}[t]
    \centering
	\includegraphics[width=\columnwidth]{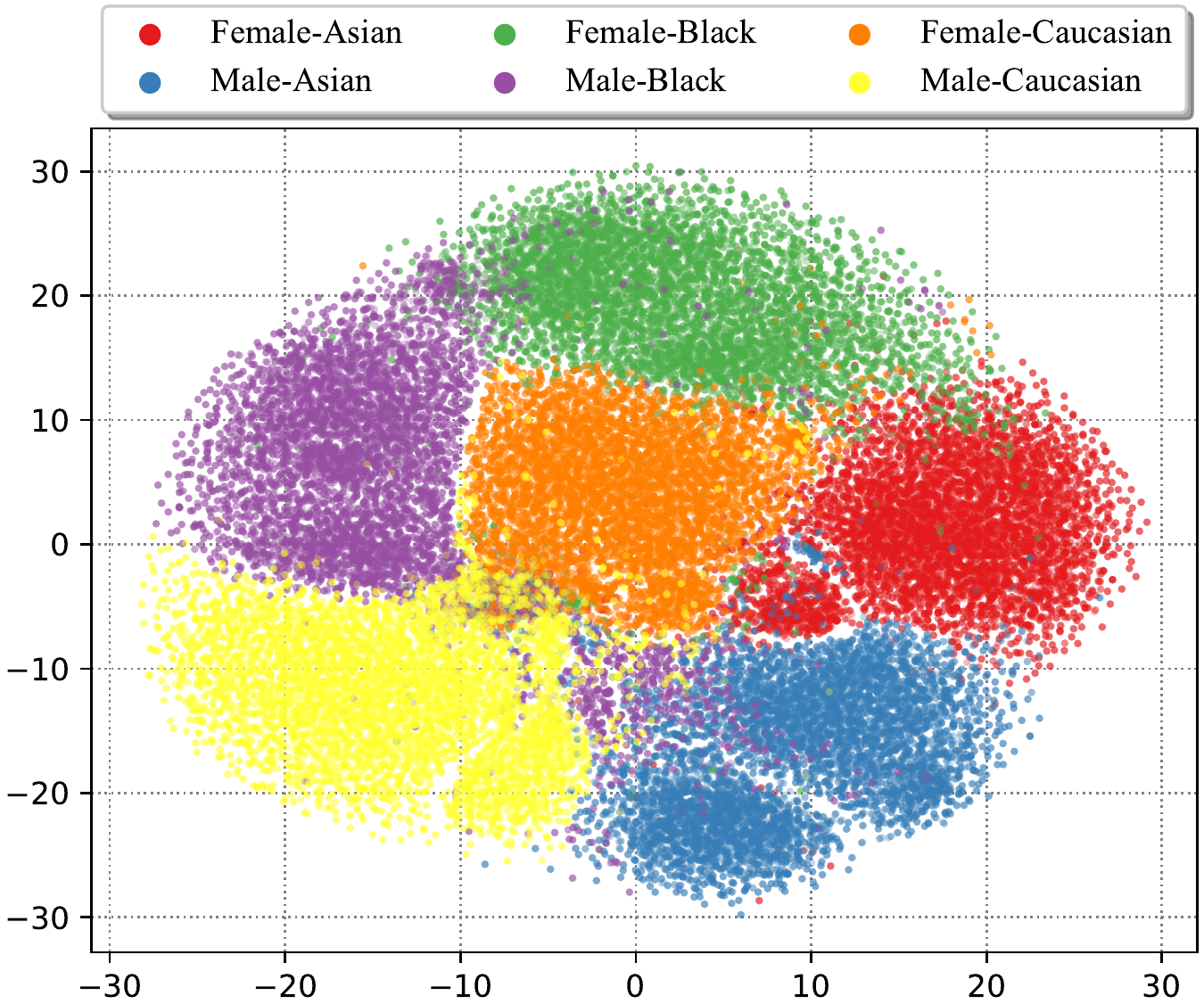}
	\caption{Data distribution of the DiveFace dataset generated by t-SNE.}
	\label{fig:diveface_tsne}
\end{figure}

DiveFace was constructed to be an unbiased face recognition dataset. It was carefully constructed with images from Megaface MF2 training dataset~\cite{kemelmacher2016megaface} that contained 4.7 million faces from 672K identities from Flickr Yahoo's dataset~\cite{thomee2015new}.

\begin{table}[]
\centering
\caption{Proportions of face images from different ethnics and genders in DiveFace dataset}
\label{table:diveface}
\resizebox{\columnwidth}{!}{
\begin{tabular}{crrr} 
\hline
\multirow{2}{*}{Ethnicity} & \multicolumn{2}{c}{Gender} & \multicolumn{1}{c}{\multirow{2}{*}{Total}} \\ 
\cline{2-3}
 & \multicolumn{1}{c}{Female} & \multicolumn{1}{c}{Male} & \multicolumn{1}{c}{} \\ 
\hline
Asian & \begin{tabular}[c]{@{}r@{}}12,000\\ (16.67\%) \end{tabular} & \begin{tabular}[c]{@{}r@{}}12,000\\ (16.67\%) \end{tabular} & \begin{tabular}[c]{@{}r@{}}24,000\\ (33.33\%) \end{tabular} \\
Black & \begin{tabular}[c]{@{}r@{}}12,000\\ (16.67\%) \end{tabular} & \begin{tabular}[c]{@{}r@{}}12,000\\ (16.67\%) \end{tabular} & \begin{tabular}[c]{@{}r@{}}24,000\\ (33.33\%) \end{tabular} \\
Caucasian & \begin{tabular}[c]{@{}r@{}}12,000\\ (16.67\%) \end{tabular} & \begin{tabular}[c]{@{}r@{}}12,000\\ (16.67\%) \end{tabular} & \begin{tabular}[c]{@{}r@{}}24,000\\ (33.33\%) \end{tabular} \\ 
\hline
Total & \begin{tabular}[c]{@{}r@{}}36,000\\ (50\%) \end{tabular} & \begin{tabular}[c]{@{}r@{}}36,000\\ (50\%) \end{tabular} & \begin{tabular}[c]{@{}r@{}}72,000\\ (100\%) \end{tabular} \\
\hline
\end{tabular}
}
\end{table}

DiveFace was designed to be evenly distributed in six demographic groups. There are 24,000 identities from six demographic groups, 4,000 identities for each group and three poses for each identity. Thus, each demographic group contained 12,000 faces for a total of 72,000 faces in the whole dataset. The DiveFace proportions for every class in the dataset are shown in Table~\ref{table:diveface}. Three ethnicity categories are available, related to the physical characteristics of each ethnic group:
\begin{itemize}
    \item Group 1: people with ancestral origin in Japan, China, Korea, and other countries in that region.
    \item Group 2: people with ancestral origins in Sub-Saharan Africa, India, Bangladesh, Bhutan, and others.
    \item Group 3: people with ancestral origins in Europe, North-America, and Latin-America with European origin.
\end{itemize}
In this study, we denoted Group 1, 2, and 3 as Asian, Black, and Caucasian, respectively. A t-distributed Stochastic Neighbor Embedding (t-SNE)~\cite{maaten2008visualizing} of dimension 2 from ResNet-50 descriptors of the full DiveFace dataset is shown in Figure~\ref{fig:diveface_tsne}. As can be seen, the six clusters separated from each other clearly. However, a few data points in the Male-Black category are also in the clusters of Male-Asian and Male-Caucasian.

\subsubsection{Labeled Faces in the Wild}
\label{sec:lfw}
Labeled Faces in the Wild (LFW) database was introduced to evaluate the performance of face verification algorithms with unconstrained parameters, such as position, pose, lighting, background, camera quality, and gender~\cite{LFWTech}. The database contains 13,233 faces collected from the web from 5,749 unique individuals.

LFW was published in 2007. It has been a very popular database in the face recognition field. LFW has already been split properly into standard training and test sets. In this work, we used the test set for evaluating the performance of our framework. It contains a balanced set of 1,000 sample pairs (500 pairs of genuine facial images and 500 pairs of imposter images).

\subsection{Experimental settings}
\label{sec:experimental_settings}
In this study, we divided the DiveFace dataset into training, validation, and test sets. The size of the training set was 60\% of the whole dataset; the size of the validation set was 10\%; and the size of the testing set was 30\%. The training set was used to train the triplet models and gender-ethnicity classifier models; the validation set was used to select optimal models; and the testing set was used to evaluate the prediction performances of all tested models. On the other hand, the performance of our full framework was evaluated with the LFW database. The average and standard deviation of the metrics of ten experimental runs, each with a different random split, are reported. For the image pairing, the set of positive samples was constructed by pairing all pose images in all possible ways within each identity. On the other hand, the set of negative samples was constructed by randomly pairing different identities.

The performance of the proposed SELM is evaluated in comparison to ResNet and ELM. ResNet is one of the most well-known DCNNs methods. As comparison baseline we used a ResNet-50 architecture pre-trained for face recognition with VGGFace2 (millions of images). The pre-trained ResNet-50 was then used to train our triplet models. These triplet models classify input image pairs into two classes (genuine or impostor match) based on Euclidean distance. The ELM and SELM methods have a similar architecture. The architecture is based on a single layer feedforward neural network that can be trained much faster than common artificial neural networks. On the other hand, SELM has one more additional layer (the Siamese layer). Both ELM and SELM use a kernel trick together with a pseudoinverse technique to generate the weights of the model that provide the lowest error rate. Moreover, we evaluate the performance when using four different types of Siamese conditions to improve the classification outcome.

As for parameter settings, the parameters of the three methods are tuned to obtain the best result. False Acceptance Rate (FAR) and False Rejection Rate (FRR) were used to find an optimal threshold, which is considered to be at the Error Equal Rate (EER). For ELM, three parameters were tuned: regularizing $C$ which was set to be $[10^{-6}, 10^{-5}, \ldots, 10^{5}, 10^{6}]$, percentage of hidden nodes which was in the range of $[10, 20, \ldots, 90, 100\%]$, and gamma $\gamma$ in RBF kernel which was $[10^{-6}, 10^{-6}]$. For SELM, two parameters were tuned: regularizing $C$ and percentage of hidden nodes. As for ResNet, we used the same Euclidean coefficients for calculating the loss function as the Euclidean coefficients that we used in the kernel trick in SELM. Hence, for the kernel trick, no parameters needed to be tuned.

\section{Results and discussion}
\label{sec:results_discussion}
In this section, we report the experimental results on the following types of evaluation: evaluation of feature performance, evaluation of classifiers, evaluation of Siamese and non-Siamese architectures, and evaluation of the performance of the whole framework. Two types of evaluation metrics are employed: verification accuracy and AUC (Area Under the Curve). For each experiment, the average and standard deviation of ten runs are reported.

\subsection{Evaluation of feature performance}
\label{sec:feature_performance}

\begin{table*}[t]
\centering
\caption{Performance metrics achieved by ResNet of each feature type on the DiveFace dataset.}
\label{table:feature_performance}
\begin{subtable}[h]{1\textwidth}
\resizebox{\textwidth}{!}{
\begin{tabular}{crrrrrrr}
\hline
\multicolumn{1}{l}{} & \multicolumn{1}{c}{Female-Asian} & \multicolumn{1}{c}{Female-Black} & \multicolumn{1}{c}{Female-Caucasian} & \multicolumn{1}{c}{Male-Asian} & \multicolumn{1}{c}{Male-Black} & \multicolumn{1}{c}{Male-Caucasian} & \multicolumn{1}{c}{Average} \\ \hline
SI & 88.8917 ± 0.40 & 92.9042 ± 0.56 & 96.5625 ± 0.34 & 90.6625 ± 0.67 & 95.4958 ± 0.42 & 97.3875 ± 0.26 & 93.6507 ± 0.44 \\
GD & 96.4375 ± 0.37 & 95.5167 ± 0.40 & 97.0667 ± 0.44 & 95.8333 ± 0.49 & 95.6333 ± 0.20 & 96.7500 ± 0.28 & 96.2063 ± 0.36 \\
GED & \textbf{98.3792 ± 0.28} & \textbf{97.2375 ± 0.33} & \textbf{98.6708 ± 0.24} & \textbf{97.7208 ± 0.27} & \textbf{97.0917 ± 0.30} & \textbf{98.1083 ± 0.16} & \textbf{97.8681 ± 0.26} \\ \hline
\end{tabular}
}
\\
\caption{Accuracy}
\label{table:feature_performance_accuracy}
\end{subtable}
\hfill
\begin{subtable}[h]{1\textwidth}
\resizebox{\textwidth}{!}{
\begin{tabular}{crrrrrrr}
\hline
\multicolumn{1}{l}{} & \multicolumn{1}{c}{Female-Asian} & \multicolumn{1}{c}{Female-Black} & \multicolumn{1}{c}{Female-Caucasian} & \multicolumn{1}{c}{Male-Asian} & \multicolumn{1}{c}{Male-Black} & \multicolumn{1}{c}{Male-Caucasian} & \multicolumn{1}{c}{Average} \\ \hline
SI & 98.2868 ± 0.40 & 96.7622 ± 0.35 & 98.4575 ± 0.25 & 98.5000 ± 0.29 & 96.9556 ± 0.24 & 98.2676 ± 0.32 & 97.8716 ± 0.31 \\
GD & 99.3332 ± 0.23 & 98.5432 ± 0.19 & 99.3619 ± 0.15 & 99.1686 ± 0.20 & 98.7926 ± 0.20 & 99.0539 ± 0.20 & 99.0422 ± 0.20 \\
GED & \textbf{99.6405 ± 0.18} & \textbf{99.0981 ± 0.28} & \textbf{99.6645 ± 0.15} & \textbf{99.5362 ± 0.14} & \textbf{99.2382 ± 0.24} & \textbf{99.5476 ± 0.11} & \textbf{99.4542 ± 0.18} \\ \hline
\end{tabular}
}
\\
\caption{AUC}
\label{table:feature_performance_auc}
\end{subtable}
\end{table*}

The performances of all features used in the experiment are presented in this section. ResNet-50 was used to train three different feature-extraction models, which were trained differently as follows.
\begin{itemize}
    \item Subject-Independent (SI) feature model: this model was trained by randomly pairing (no pattern) individuals as training samples, e.g., no pre-assigned values for proportions of gender and ethnicity classes. This kind of model training is conventional in face recognition.
    \item Gender-Dependent (GD) feature model: this model was trained independently for Males and Females.
    \item Gender-Ethnicity-Dependent (GED) feature model: this model focused on facial characteristics of each cohort, thus the model was trained independently on each of the six considered cohorts. The number of training samples from every cohort was assigned to be the same.
\end{itemize}
The experimental results on the DiveFace dataset are shown in Table~\ref{table:feature_performance_accuracy} and \ref{table:feature_performance_auc}. The best features among all types of features in every cohort are marked in bold.

As can be seen in Table~\ref{table:feature_performance}, the values of Accuracy and AUC reflect each other, the higher the Accuracy, the higher the AUC, and vice versa. The feature performance of SI, the baseline, was the worst, but it still reached up to 93.65\% and 97.87\% in overall accuracy and AUC, respectively. Therefore, it was a challenge to improve on those metrics. Nevertheless, GED and GD were able to yield a better AUC performance: 99.45\% and 99.04\% AUC value, respectively. GED results to be the best among the tested methods, followed by GD. Furthermore, compared to SI and GD, GED exhibits better metrics for every cohort. This result confirms our hypothesis that training samples with specific, distinctive groups could induce the model to learn more useful facial features. The reason that the performance of GD was higher than SI and that the performance of GED was higher than GD is that GD learned intensively and independently on gender group, and GED learned in the same way as GD but on both gender and ethnicity groups.

Nevertheless, GED performance was only $0.41\%$ better than that of GD. To check if that difference was significant or not, we used one-way ANOVA to test the null hypothesis (SI, GD, and GED have the same population mean, $\mu_{SI}=\mu_{GD}=\mu_{GED}$)~\cite{siegel1956nonparametric}. The statistical result, $f=144.06$, indicates that the difference is statistically significant at a level of $p<0.001$, hence the null hypothesis $H_0$ was rejected. GED is the best feature type among the three models tested in this work.

\subsection{Evaluation of classifier performance}
\label{sec:method_performance}

\begin{table*}[t]
\centering
\caption{Performance metrics on the DiveFace dataset achieved by the proposed SELM in comparison to standard ELM and the ResNet baseline using the most robust feature (GED).}

\label{table:alg_performance}
\begin{subtable}[h]{1\textwidth}
\resizebox{\textwidth}{!}{
\begin{tabular}{crrrrrrr}
\hline
 & \multicolumn{1}{c}{Female-Asian} & \multicolumn{1}{c}{Female-Black} & \multicolumn{1}{c}{Female-Caucasian} & \multicolumn{1}{c}{Male-Asian} & \multicolumn{1}{c}{Male-Black} & \multicolumn{1}{c}{Male-Caucasian} & \multicolumn{1}{c}{Average} \\ \hline
ResNet & 98.3792 ± 0.28 & 97.2375 ± 0.33 & 98.6708 ± 0.24 & 97.7208 ± 0.27 & 97.0917 ± 0.30 & 98.1083 ± 0.16 & 97.8681 ± 0.26 \\
ELM & 97.8917 ± 0.32 & 96.9000 ± 0.36 & 98.0667 ± 0.35 & 96.0583 ± 0.56 & 97.0750 ± 0.36 & 98.0000 ± 0.41 & 97.3319 ± 0.39 \\
SELM\textsubscript{Sum} & 98.7042 ± 0.27 & 97.7625 ± 0.58 & \textbf{98.9083 ± 0.23} & 97.8917 ± 0.37 & \textbf{98.0042 ± 0.33} & 98.6167 ± 0.22 & 98.3146 ± 0.33 \\
SELM\textsubscript{Dist} & 97.6125 ± 1.33 & 97.5208 ± 0.55 & 98.8208 ± 0.33 & 98.0333 ± 0.17 & 97.6833 ± 0.42 & \textbf{98.6250 ± 0.19} & 98.0493 ± 0.50 \\
SELM\textsubscript{Mult} & 98.5125 ± 0.32 & 97.4750 ± 0.60 & 98.7333 ± 0.30 & \textbf{98.3750 ± 0.28} & 97.6208 ± 0.36 & 98.4708 ± 0.20 & 98.1979 ± 0.34 \\
SELM\textsubscript{Mean} & \textbf{98.7125 ± 0.28} & \textbf{97.7708 ± 0.58} & \textbf{98.9083 ± 0.23} & 97.9083 ± 0.34 & \textbf{98.0042 ± 0.33} & 98.6208 ± 0.22 & \textbf{98.3208 ± 0.33} \\ \hline
\end{tabular}
}
\\
\caption{Accuracy}
\label{table:alg_performance_accuracy}
\end{subtable}
\hfill
\begin{subtable}[h]{1\textwidth}
\resizebox{\textwidth}{!}{
\begin{tabular}{crrrrrrr}
\hline
 & \multicolumn{1}{c}{Female-Asian} & \multicolumn{1}{c}{Female-Black} & \multicolumn{1}{c}{Female-Caucasian} & \multicolumn{1}{c}{Male-Asian} & \multicolumn{1}{c}{Male-Black} & \multicolumn{1}{c}{Male-Caucasian} & \multicolumn{1}{c}{Average} \\ \hline
ResNet & 99.6405 ± 0.18 & 99.0981 ± 0.28 & 99.6645 ± 0.15 & 99.5362 ± 0.14 & 99.2382 ± 0.24 & 99.5476 ± 0.11 & 99.4542 ± 0.18 \\
ELM & 99.6763 ± 0.15 & 99.2188 ± 0.23 & 99.7290 ± 0.11 & 99.5469 ± 0.14 & 99.4825 ± 0.14 & 99.7391 ± 0.09 & 99.5654 ± 0.14 \\
SELM\textsubscript{Sum} & 99.7911 ± 0.09 & \textbf{99.4315 ± 0.20} & 99.8274 ± 0.10 & 99.7653 ± 0.09 & 99.6747 ± 0.11 & 99.8497 ± 0.06 & \textbf{99.7233 ± 0.11} \\
SELM\textsubscript{Dist} & 99.7322 ± 0.11 & 99.3970 ± 0.20 & 99.7763 ± 0.10 & 99.7031 ± 0.11 & 99.6437 ± 0.13 & 99.8311 ± 0.05 & 99.6806 ± 0.12 \\
SELM\textsubscript{Mult} & 99.6263 ± 0.19 & 99.1449 ± 0.28 & 99.6647 ± 0.15 & \textbf{99.7915 ± 0.08} & 99.3344 ± 0.23 & 99.5760 ± 0.12 & 99.5230 ± 0.17 \\
SELM\textsubscript{Mean} & \textbf{99.7913 ± 0.09} & 99.4286 ± 0.20 & \textbf{99.8280 ± 0.10} & 99.7659 ± 0.09 & \textbf{99.6751 ± 0.11} & \textbf{99.8503 ± 0.06} & 99.7232 ± 0.11 \\ \hline
\end{tabular}
}
\\
\caption{AUC}
\label{table:alg_performance_auc}
\end{subtable}
\end{table*}

\begin{figure}[t]
    \centering
	\includegraphics[width=\columnwidth]{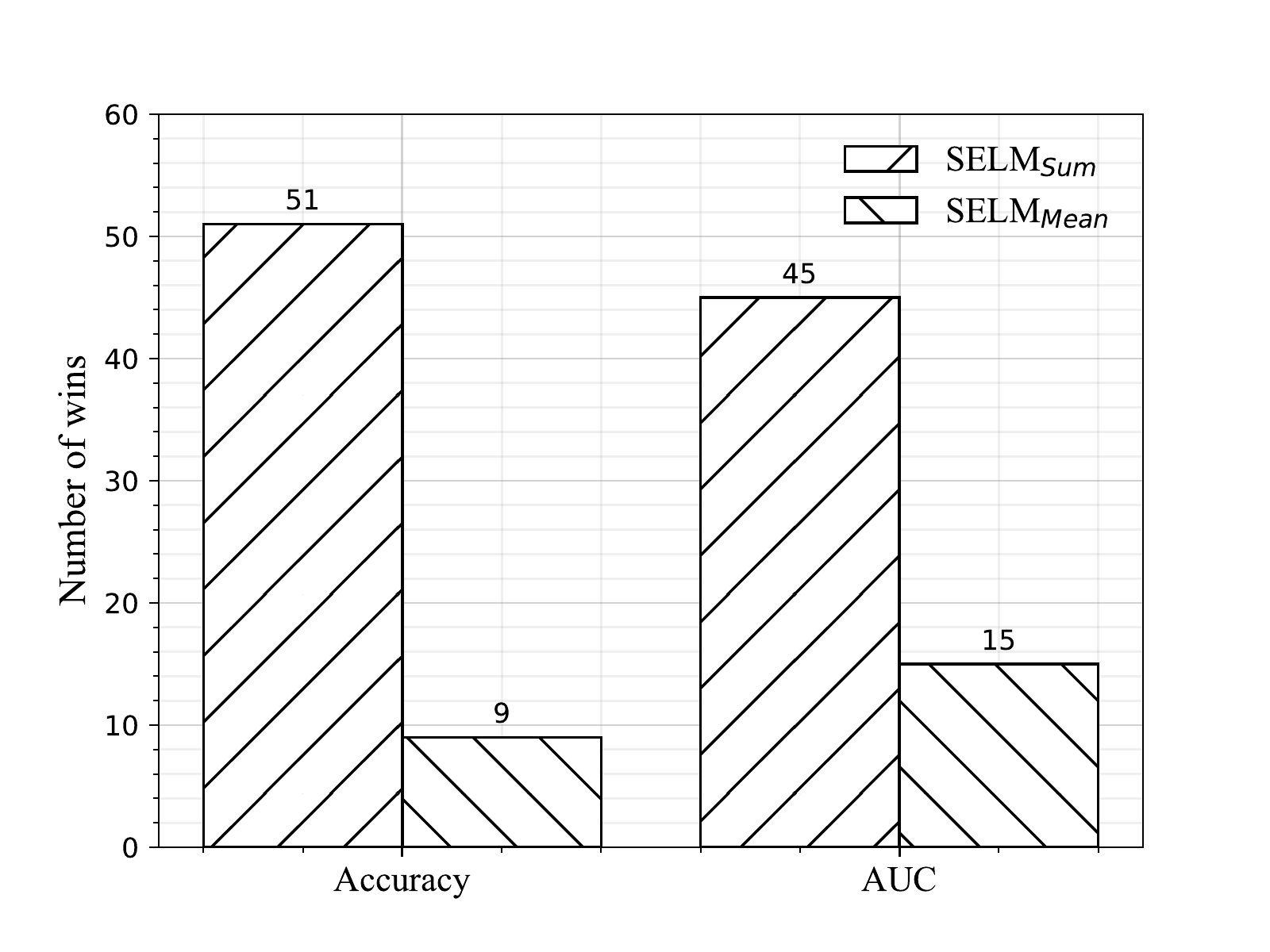}
	\caption{Comparison of number of wins accomplished by SELM\textsubscript{Sum} and SELM\textsubscript{Mean} in terms of Accuracy and AUC evaluation metrics.}
	\label{fig:compare_overcome}
\end{figure}

\begin{figure}[t]
    \centering
	\includegraphics[width=\columnwidth]{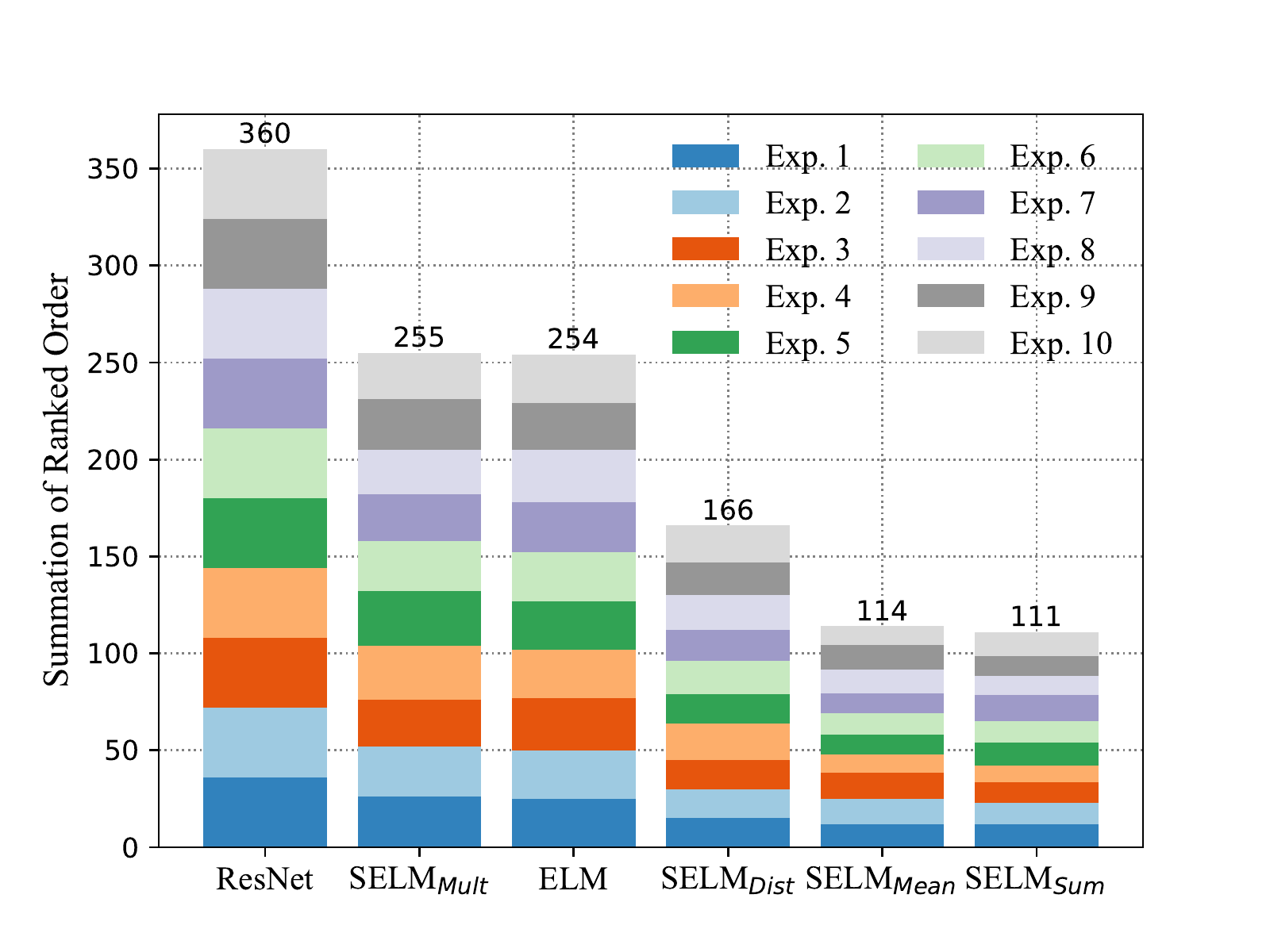}
	\caption{Summation of ranked order in terms of AUC performance, reported as stacked bars in descending order of ten experiments.}
	\label{fig:rankedorder}
\end{figure}

The performances of ResNet, ELM, and SELM embedded with four different types of Siamese conditions---summation, distance, multiply, and mean denoted as Sum, Dist, Mult, and Mean, respectively---are shown in Table~\ref{table:alg_performance_accuracy} and \ref{table:alg_performance_auc}. We used the best feature, GED, obtained from the previous experiment, Section~\ref{sec:feature_performance}. Table~\ref{table:alg_performance} lists the performance metrics---Accuracy and AUC---achieved by the proposed SELM in comparison to standard ELM and the ResNet baseline. The best metric achieved by the best classifier candidate for each identity cohort is marked in bold.

The experimental results in Table~\ref{table:alg_performance_accuracy} show that SELM\textsubscript{Mean} is the best classification method in terms of overall accuracy score, followed by SELM\textsubscript{Sum}, SELM\textsubscript{Mult}, SELM\textsubscript{Dist}, ResNet, and ELM. SELM\textsubscript{Mean} yields the highest accuracy for four out of the six demographic groups; SELM\textsubscript{Sum} yields the highest accuracy for two out of the six demographic groups; and SELM\textsubscript{Mult} yields the highest accuracy for one group. Nevertheless, the accuracy score achieved by the first and second best methods, SELM\textsubscript{Mean} and SELM\textsubscript{Sum}, differs only by $0.01\%$. Furthermore, SELM\textsubscript{Sum} achieves the highest AUC metric ($99.72$) for only one out of six groups, but SELM\textsubscript{Mean}, ($99.72$) achieves the highest AUC for four out of the six groups. SELM\textsubscript{Dist}, ELM, SELM\textsubscript{Mult} and ResNet follow those two in this order.  Figure~\ref{fig:compare_overcome} shows a comparison between the number of wins of SELM\textsubscript{Sum} and SELM\textsubscript{Mean}, in terms of both Accuracy and AUC evaluation metrics. Since the graphs were data from ten experimental runs of six demographic cohorts, the ideal score should be $10 \times 6 = 60$. Figure~\ref{fig:compare_overcome} shows clearly that SELM\textsubscript{Sum} is definitely better than SELM\textsubscript{Mean} for 51 out of 60 cases in terms of accuracy and 45 out of 60 cases in terms of AUC.

In addition, as a way to rank the methods, we show in Figure~\ref{fig:rankedorder} the accumulated AUC score ranks across the ten experimental runs. The ideal summation would be 1\textsuperscript{st} rank in all 60 experimental runs, i.e., 60 is the lowest summation possible (best method). At the other extreme, 6\textsuperscript{th}, 360 would be the highest accumulated rank possible (worst method). We then used Kendall's Coefficient of Concordance~$\mathcal{W}$, a statistical technique, to calculate the degree of reliability of the ranked order:

\begin{equation}
\mathcal{W}=\frac{12\sum_{i=1}^{N} \bar{R}_{i}^{2}-3N(N+1)^{2}}{N(N^{2}-1)},
\end{equation}

where $\bar{R}$ is the average ranked order assigned to the $i$-th candidate; $N$ is the number of candidate methods (six); and the number of runs times the number of cohort groups $k$ is 60. The value of $\mathcal{W}$ was found to be $0.7526$. The critical value in $\chi^{2}$ distribution was converted from $\mathcal{W}$ by the following equation:

\begin{equation}
\chi^{2}= k(N-1)\mathcal{W}.
\end{equation}

We acquired $\chi^{2}=225.78$ which indicates that the ranked order shown in Figure~\ref{fig:rankedorder} is reliable at a confidence level of 99.9\%. The rank order is as follows:

\begin{align*}
SELM\textsubscript{Sum}>SELM\textsubscript{Mean}>SELM\textsubscript{Dist}>\\
ELM>SELM\textsubscript{Mult}>ResNet
\end{align*}

\subsection{Evaluation of Siamese and non-Siamese architectures performance}

\begin{figure*}[t]
\centering
\begin{subfigure}[b]{\columnwidth}
\includegraphics[width=\columnwidth]{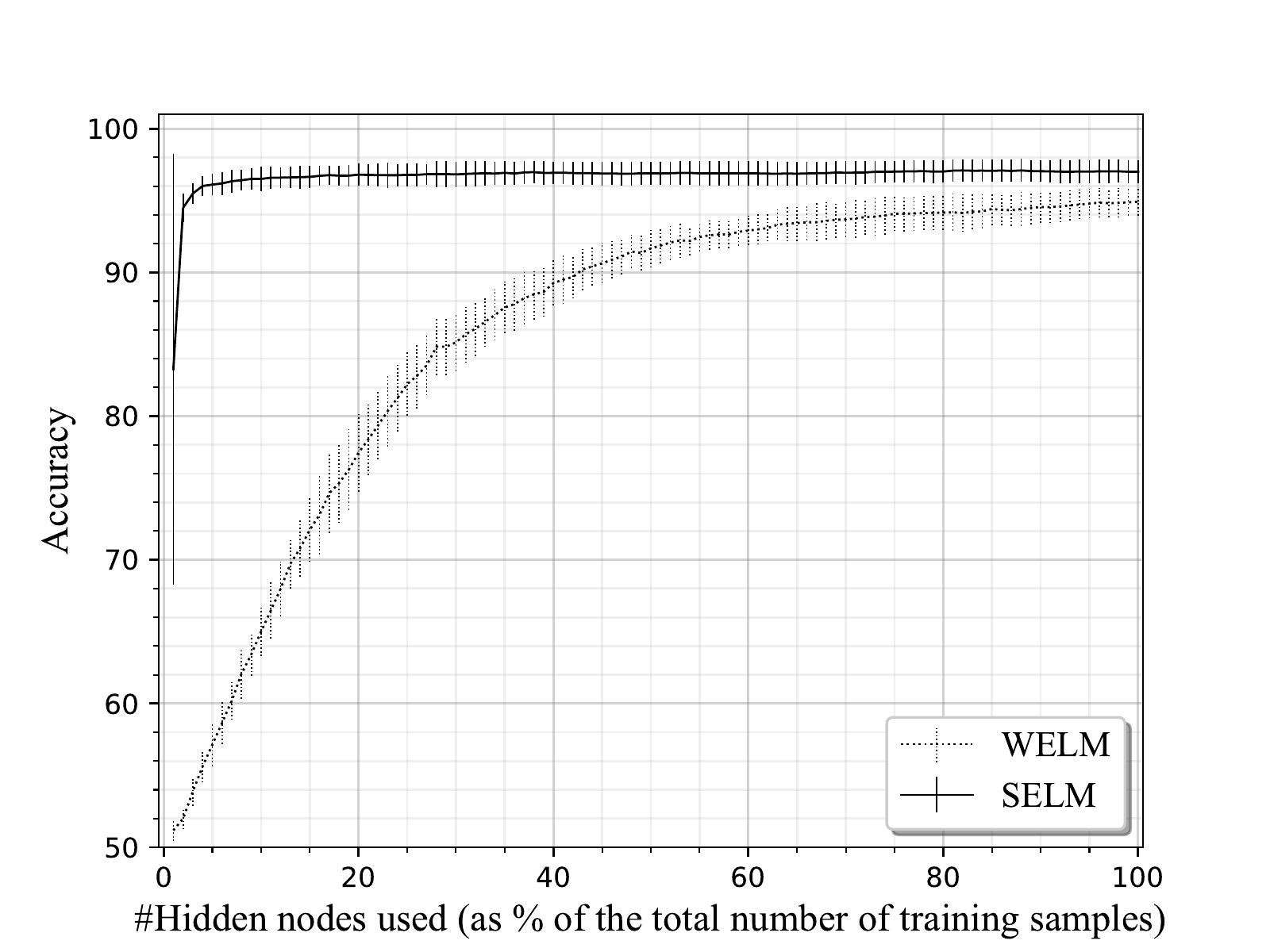}
\caption{Accuracy}
\label{fig:varynode_accuracy}
\end{subfigure}
\hfill
\begin{subfigure}[b]{\columnwidth}
\includegraphics[width=\columnwidth]{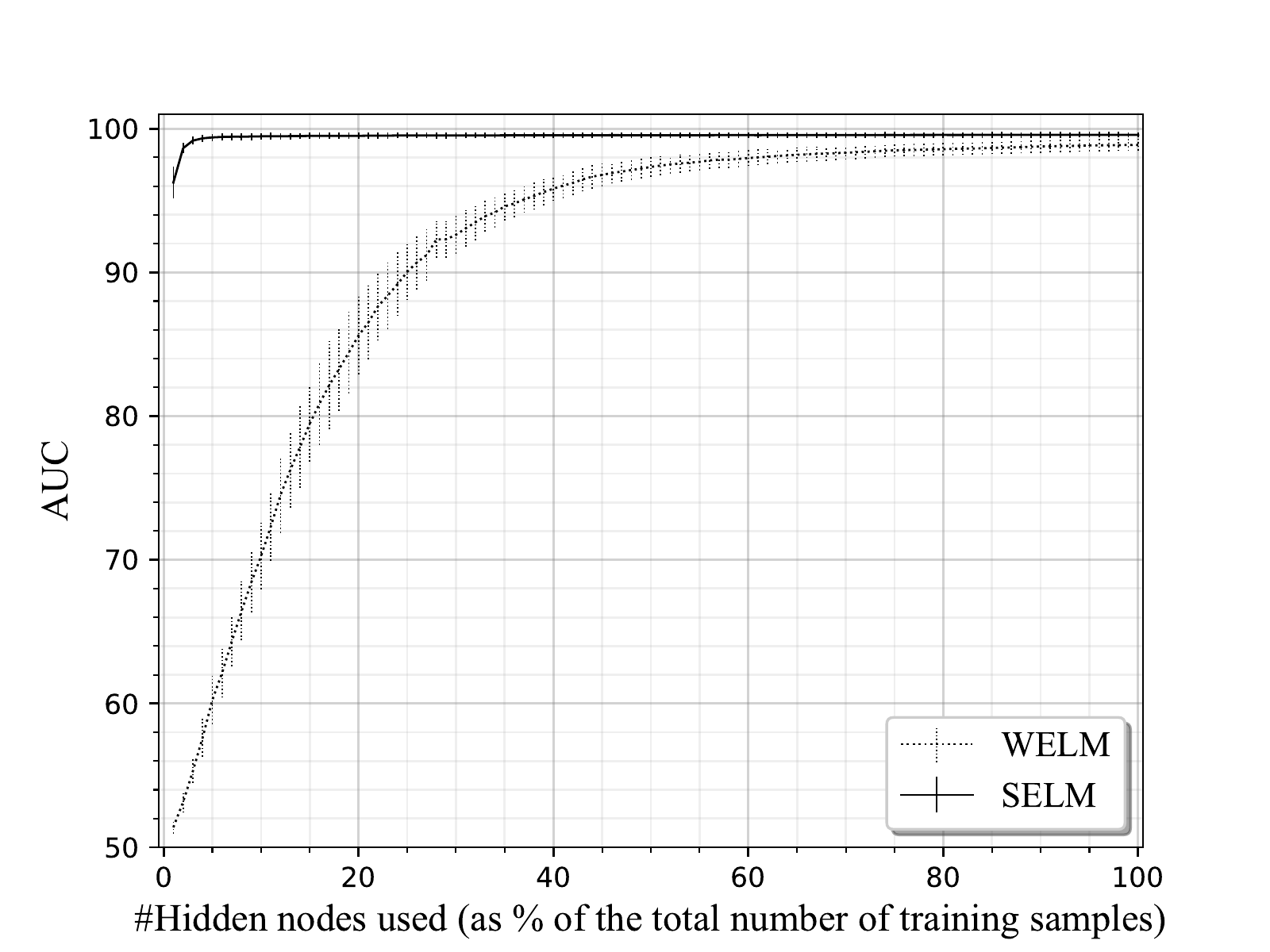}
\caption{AUC}
\label{fig:varynode_auc}
\end{subfigure}
\caption{The performances of Siamese (SELM) VS non-Siamese Extreme Learning Machines (WELM).}
\label{fig:varynode}
\end{figure*}

In this section, we compare the performance of the most robust Siamese architecture (SELM\textsubscript{Sum}) to that of WELM, an ELM with non-Siamese architecture. Their backbone architecture was identical except the additional Siamese layer in SELM. Simultaneous dual inputs into WELM were concatenated for training the network, but these inputs were not concatenated by SELM; instead, they were passed through the Siamese layer. Any subsequent procedural steps of the two architectures are the same.

Figure~\ref{fig:varynode} shows the Accuracy and AUC performances of WELM and SELM while using an incresing number of hidden nodes to train a model (Figure~\ref{fig:varynode_accuracy} and \ref{fig:varynode_auc}, respectively). The performance values are obtained averaging across the six available demographic cohorts. It can be seen that WELM has to use a large number of hidden nodes up to $80\%$ of the training samples in order to compete with SELM, while SELM needs only less than $10\%$ in order to achieve excellent results. The optimal model of WELM achieves $94\%$ Accuracy when the number of its hidden nodes is $99.0\%$, while SELM achieves $97.00\%$ Accuracy with a number of hidden nodes of only $81\%$. It should also be noted that SELM was able to achieve $96.80\%$ Accuracy and $99.50\%$ AUC with a number of hidden nodes of only $20\%$. We used two-sample $t$-test analysis to check the statistical significance between the mean scores from both methods at $p<0.001$ and found that the t-values for Accuracy and AUC are $t=9.08$ and $t=6.78$, respectively. Hence, we conclude that the proposed Siamese-ELM performs significantly better than the standard non-Siamese-ELM.

\subsection{Evaluation of the whole system performance}

\begin{figure}[t]
\centering
\begin{subfigure}[b]{\columnwidth}
\includegraphics[width=\columnwidth]{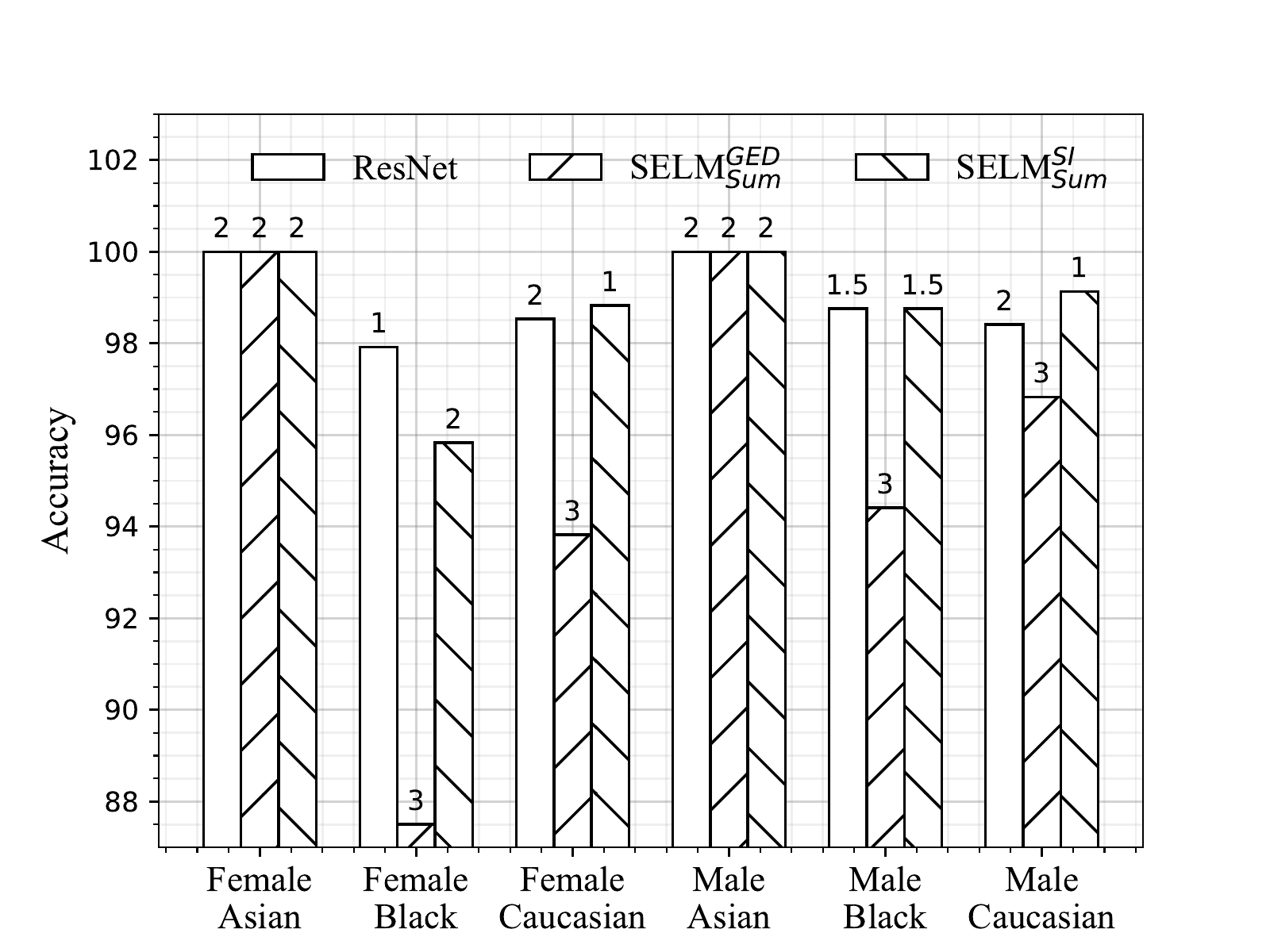}
\end{subfigure}
\caption{System’s face verification Accuracy and ranked order for each demographic group of the LFW database.}
\label{fig:framework_perf}
\end{figure}

\begin{figure}[t]
\centering
\begin{subfigure}[b]{\columnwidth}
\includegraphics[width=\columnwidth]{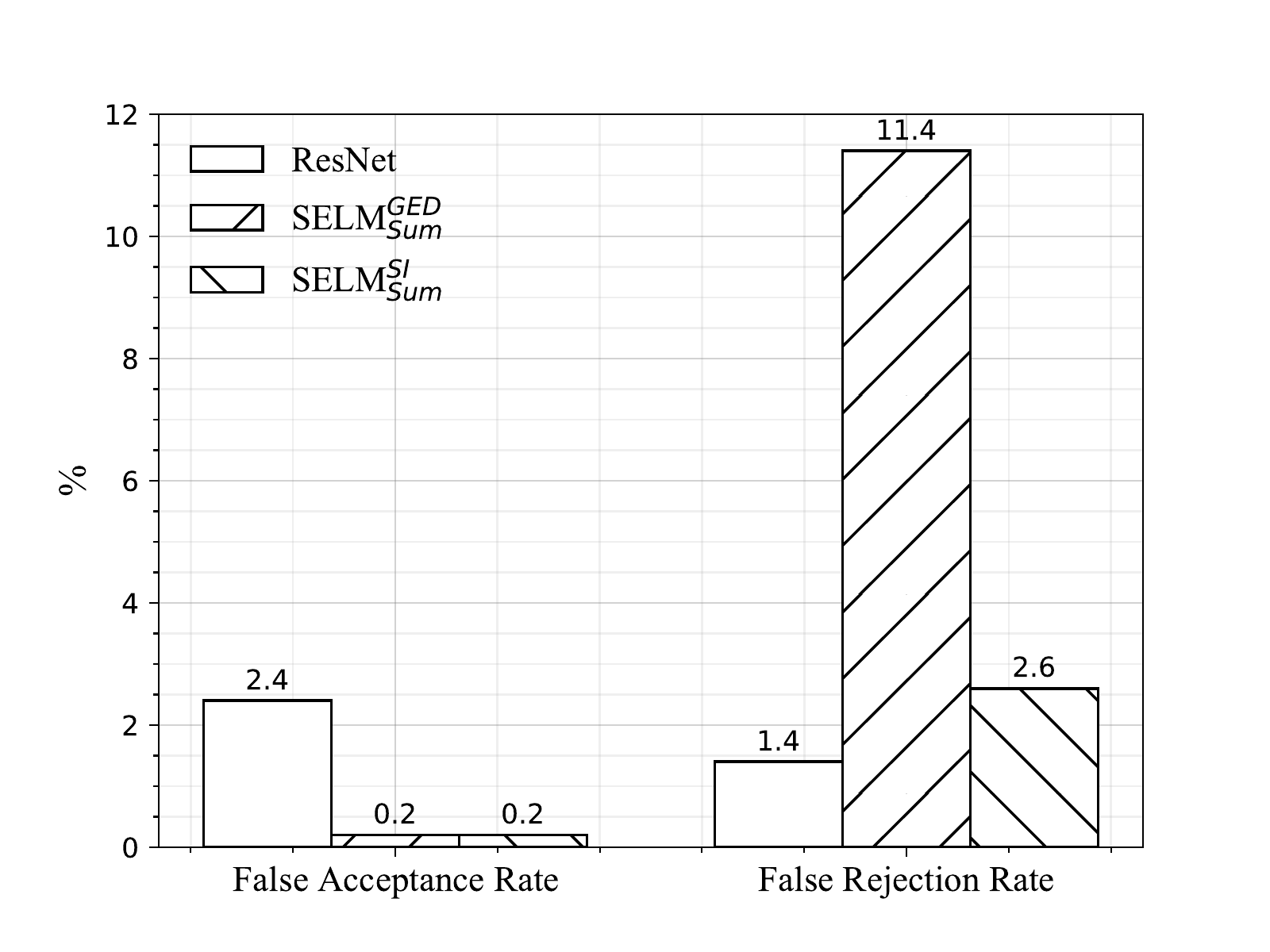}
\end{subfigure}
\caption{False Acceptance Rate and False Rejection Rate of the proposed SELM systems for the LFW database.}
\label{fig:framework_conmat}
\end{figure}


We evaluated the proposed system, described in Section~\ref{sec:proposed_framework}, in conjunction with the most robust feature, GED, described in Section~\ref{sec:feature_performance}, and the most robust classification method, SELM\textsubscript{Sum}, described in Section~\ref{sec:method_performance}. The whole system is termed SELM\textsupsub{GED}{Sum}. It should be noted that the proposed system first classifies individuals according to their respective Gender-Ethnicity class so that a proper feature-extraction model could be selected for that purpose, and the input image pairs that are not in the same Gender-Ethnicity class are classified as impostor comparisons. SELM\textsupsub{SI}{Sum} is similar to SELM\textsupsub{GED}{Sum} but without the initial Gender-Ethnicity classification. In Figure~\ref{fig:framework_perf} we show the performances of ResNet (baseline), SELM\textsupsub{GED}{Sum}, and SELM\textsupsub{SI}{Sum} tested on the standard test set of the LFW database.

The ranked order of each demographic is shown on the top of the bar representing that group in Figure~\ref{fig:framework_perf}. It can be seen that SELM\textsupsub{SI}{Sum} is the best method producing the smallest sum of ranked order ($9.5$), followed by ResNet  ($10.5$), and SELM\textsupsub{GED}{Sum} ($16$). The performances of both SELM\textsupsub{SI}{Sum} and SELM\textsupsub{GED}{Sum} for the Black demographic class are lower than the performance obtained for Asian and Caucasian classes. This is because the systems were trained on DiveFace, which contains data of individuals in the Black group whose origin is in the Sub-Saharan region, Africa, India, Bangladesh, and Bhutan, while the tested LFW dataset contains data of individuals of the Black group not well represented by those regions. Regarding the performance of SELM\textsupsub{GED}{Sum}, it works like a two-stage prediction system, and the accuracy of the final prediction in the second stage depends highly on the performance of the first stage, the Gender-Ethnicity prediction model. In this study, SELM\textsupsub{GED}{Sum} yielded very accurate outcomes when the first stage provided an ideal classification of Gender-Ethnicity group. Figure~\ref{fig:framework_conmat} shows bar graphs of two evaluation metrics---False Acceptance Rate (FAR) and False Rejection Rate (FRR)---produced by ResNet, SELM\textsupsub{GED}{Sum}, and SELM\textsupsub{SI}{Sum}. FAR was considered the most important metric for this kind of task. It represented the rate of which wrong persons were given access to the system. The performance results show that both SELM\textsupsub{GED}{Sum} and SELM\textsupsub{SI}{Sum} provided a very low FAR ($0.2\%$), 12 times lower than that provided by ResNet ($2.4\%$), indicating that they would execute with far less error in face recognition tasks.

\section{Conclusion}
\label{sec:conclusion}
A framework for face verification is proposed. The framework employs a new classification method called Siamese Extreme Learning Machine (SELM), an improved version of a powerful classification method called Extreme Learning Machine that can accept two image inputs in parallel and process them concurrently. In our performance evaluation, SELM was studied in conjunction with several features that were trained on unbiased demographic-dependent groups. With this training, the feature-extraction model in our proposed SELM was able to better recognize distinct features of individuals in demographic groups than a conventional feature-extraction model was able to. In an evaluation experiment, four different types of Siamese conditions embedded in the Siamese layer were compared. The SELM with summation and mean conditions provided the highest overall performance score. Furthermore, in another experiment, SELM with ‘sum’ Siamese condition was demonstrated to be more robust than baseline methods ResNet and ELM. In particular, the proposed method was able to perform the verification task better, with $98.31\%$ Accuracy and $99.72\%$ AUC, than the other methods. More importantly, SELM\textsupsub{SI}{Sum} provided a very low $0.2\%$ false acceptance rate, which was $12$ times lower than that provided by ResNet ($2.4\%$), a considerable improvement.

For future work, we aim to do the following: (i) train our own face recognition model from scratch to eliminate any bias from the beginning \cite{ter21bias}, (ii) explore other architectures for processing multiple inputs on top of ELM backbones beyond Siamese settings using recent advances from the information fusion field \cite{fierrez18fusion2}, and (iii) applying SELM to other types of image comparison tasks in addition to human face verification.

%
%
\section*{Declarations}
\paragraph{Funding} 
This work was supported by the Faculty of Information Technology, King Mongkut's Institute of Technology Ladkrabang and project BIBECA (RTI2018-101248-B-I00 MINECO/FEDER). 

\paragraph{Conflicts of interest/Competing interests}
The authors declare that they have no competing interests.


\bibliographystyle{spmpsci} 
\bibliography{references.bib}

\begin{thebibliography}{10}
\providecommand{\url}[1]{{#1}}
\providecommand{\urlprefix}{URL }
\expandafter\ifx\csname urlstyle\endcsname\relax
  \providecommand{\doi}[1]{DOI~\discretionary{}{}{}#1}\else
  \providecommand{\doi}{DOI~\discretionary{}{}{}\begingroup
  \urlstyle{rm}\Url}\fi

\bibitem{acien18bias}
Acien, A., Morales, A., Vera-Rodriguez, R., Bartolome, I., Fierrez, J.:
  Measuring the gender and ethnicity bias in deep models for face recognition.
\newblock In: IAPR Iberoamerican Congress on Pattern Recognition (CIARP),
  \emph{LNCS}, vol. 11401, pp. 584--593. Springer (2018)

\bibitem{alonso09finger}
Alonso-Fernandez, F., Bigun, J., Fierrez, J., Fronthaler, H., Kollreider, K.,
  Ortega-Garcia, J.: Fingerprint Recognition.
\newblock Springer-Verlag, London (2009).
\newblock ISBN 978-1-84800-291-3

\bibitem{antipov2015learned}
Antipov, G., Berrani, S.A., Ruchaud, N., Dugelay, J.L.: Learned vs.
  hand-crafted features for pedestrian gender recognition.
\newblock In: Proceedings of the 23rd ACM International Conference on
  Multimedia, pp. 1263--1266 (2015)

\bibitem{arca2003face}
Arca, S., Campadelli, P., Lanzarotti, R.: A face recognition system based on
  local feature analysis.
\newblock In: International Conference on Audio and Video-based Biometric
  Person Authentication, pp. 182--189. Springer (2003)

\bibitem{bianco2017large}
Bianco, S.: Large age-gap face verification by feature injection in deep
  networks.
\newblock Pattern Recognition Letters \textbf{90}, 36--42 (2017)

\bibitem{buolamwini2018gender}
Buolamwini, J., Gebru, T.: Gender shades: Intersectional accuracy disparities
  in commercial gender classification.
\newblock In: Conference on Fairness, Accountability and Transparency, pp.
  77--91 (2018)

\bibitem{cao2018vggface2}
Cao, Q., Shen, L., Xie, W., Parkhi, O.M., Zisserman, A.: Vggface2: A dataset
  for recognising faces across pose and age.
\newblock In: 2018 13th IEEE International Conference on Automatic Face \&
  Gesture Recognition (FG 2018), pp. 67--74. IEEE (2018)

\bibitem{chen2012dictionary}
Chen, Y.C., Patel, V.M., Phillips, P.J., Chellappa, R.: Dictionary-based face
  recognition from video.
\newblock In: European Conference on Computer vision, pp. 766--779. Springer
  (2012)

\bibitem{cook2019demographic}
Cook, C.M., Howard, J.J., Sirotin, Y.B., Tipton, J.L., Vemury, A.R.:
  Demographic effects in facial recognition and their dependence on image
  acquisition: An evaluation of eleven commercial systems.
\newblock IEEE Transactions on Biometrics, Behavior, and Identity Science
  \textbf{1}(1), 32--41 (2019)

\bibitem{dadi2016improved}
Dadi, H.S., Pillutla, G.M.: Improved face recognition rate using hog features
  and svm classifier.
\newblock IOSR Journal of Electronics and Communication Engineering
  \textbf{11}(4), 34--44 (2016)

\bibitem{del2018introducing}
Del~Sole, A.: Introducing microsoft cognitive services.
\newblock In: Microsoft Computer Vision APIs Distilled, pp. 1--4. Springer
  (2018)

\bibitem{arcFace}
Deng, J., Guo, J., Niannan, X., Zafeiriou, S.: Arcface: Additive angular margin
  loss for deep face recognition.
\newblock In: Proc. IEEE Conference on Computer Vision and Pattern Recognition
  (2019)

\bibitem{Deng_2020_CVPR}
Deng, J., Guo, J., Ververas, E., Kotsia, I., Zafeiriou, S.: Retinaface:
  Single-shot multi-level face localisation in the wild.
\newblock In: Proceedings of the IEEE/CVF Conference on Computer Vision and
  Pattern Recognition (CVPR) (2020)

\bibitem{fierrez06phd}
Fierrez, J.: Adapted Fusion Schemes for Multimodal Biometric Authentication.
\newblock PhD Thesis, Univ. Politecnica de Madrid, Spain (2006)

\bibitem{fierrez18fusion}
Fierrez, J., Morales, A., Vera-Rodriguez, R., Camacho, D.: Multiple classifiers
  in biometrics. part 1: Fundamentals and review.
\newblock Information Fusion \textbf{44}, 57--64 (2018).
\newblock \doi{https://doi.org/10.1016/j.inffus.2017.12.003}

\bibitem{fierrez18fusion2}
Fierrez, J., Morales, A., Vera-Rodriguez, R., Camacho, D.: Multiple classifiers
  in biometrics. part 2: Trends and challenges.
\newblock Information Fusion \textbf{44}, 103--112 (2018).
\newblock \doi{https://doi.org/10.1016/j.inffus.2017.12.005}

\bibitem{fierrez18touch}
Fierrez, J., Pozo, A., Martinez-Diaz, M., Galbally, J., Morales, A.:
  Benchmarking touchscreen biometrics for mobile authentication.
\newblock IEEE Trans. on Information Forensics and Security \textbf{13}(11),
  2720--2733 (2018).
\newblock \doi{https://doi.org/10.1109/TIFS.2018.2833042}

\bibitem{galbally2019study}
Galbally, J., Ferrara, P., Haraksim, R., Psyllos, A., Beslay, L.: Study on face
  identification technology for its implementation in the schengen information
  system.
\newblock Publications Office of the European Union  (2019)

\bibitem{sosa18cots}
Gonzalez-Sosa, E., Fierrez, J., Vera-Rodriguez, R., Alonso-Fernandez, F.:
  Facial soft biometrics for recognition in the wild: Recent works, annotation
  and {COTS} evaluation.
\newblock IEEE Trans. on Information Forensics and Security \textbf{13}(8),
  2001--2014 (2018).
\newblock \doi{https://doi.org/10.1109/TIFS.2018.2807791}

\bibitem{goswami2017face}
Goswami, G., Vatsa, M., Singh, R.: Face verification via learned representation
  on feature-rich video frames.
\newblock IEEE Transactions on Information Forensics and Security
  \textbf{12}(7), 1686--1698 (2017)

\bibitem{guo2020learning}
Guo, J., Zhu, X., Zhao, C., Cao, D., Lei, Z., Li, S.Z.: Learning meta face
  recognition in unseen domains.
\newblock In: Proceedings of the IEEE/CVF Conference on Computer Vision and
  Pattern Recognition, pp. 6163--6172 (2020)

\bibitem{gurpinar2016kernel}
Gurpinar, F., Kaya, H., Dibeklioglu, H., Salah, A.: Kernel elm and cnn based
  facial age estimation.
\newblock In: Proceedings of the IEEE Conference on Computer Vision and Pattern
  Recognition Workshops, pp. 80--86 (2016)

\bibitem{fierrez21faceq}
Hernandez-Ortega, J., Galbally, J., Fierrez, J., Beslay, L.: Biometric quality:
  Review and application to face recognition with {FaceQnet}.
\newblock arXiv:2006.03298  (2021)

\bibitem{high2012era}
High, R.: The era of cognitive systems: An inside look at ibm watson and how it
  works.
\newblock IBM Corporation, Redbooks pp. 1--16 (2012)

\bibitem{hoffer2015deep}
Hoffer, E., Ailon, N.: Deep metric learning using triplet network.
\newblock In: International Workshop on Similarity-based Pattern Recognition,
  pp. 84--92. Springer (2015)

\bibitem{LFWTech}
Huang, G.B., Ramesh, M., Berg, T., Learned-Miller, E.: Labeled faces in the
  wild: A database for studying face recognition in unconstrained environments.
\newblock Tech. Rep. 07-49, University of Massachusetts, Amherst (2007)

\bibitem{huang2004extreme}
Huang, G.B., Zhu, Q.Y., Siew, C.K.: Extreme learning machine: a new learning
  scheme of feedforward neural networks.
\newblock In: 2004 IEEE International Joint Conference on Neural Networks (IEEE
  Cat. No. 04CH37541), vol.~2, pp. 985--990. IEEE (2004)

\bibitem{jain16years}
Jain, A.K., Nandakumar, K., Ross, A.: 50 years of biometric research:
  Accomplishments, challenges, and opportunities.
\newblock Pattern Recognition Letters \textbf{79}, 80--105 (2016).
\newblock \doi{https://doi.org/10.1016/j.patrec.2015.12.013}

\bibitem{jin2014hand}
Jin, L., Gao, S., Li, Z., Tang, J.: Hand-crafted features or machine learnt
  features? together they improve rgb-d object recognition.
\newblock In: 2014 IEEE International Symposium on Multimedia, pp. 311--319.
  IEEE (2014)

\bibitem{kemelmacher2016megaface}
Kemelmacher-Shlizerman, I., Seitz, S.M., Miller, D., Brossard, E.: The megaface
  benchmark: 1 million faces for recognition at scale.
\newblock In: Proceedings of the IEEE conference on computer Vision and Pattern
  Recognition, pp. 4873--4882 (2016)

\bibitem{klare2012face}
Klare, B.F., Burge, M.J., Klontz, J.C., Bruegge, R.W.V., Jain, A.K.: Face
  recognition performance: Role of demographic information.
\newblock IEEE Transactions on Information Forensics and Security
  \textbf{7}(6), 1789--1801 (2012)

\bibitem{kudisthalert2020counting}
Kudisthalert, W., Pasupa, K., Tongsima, S.: Counting and classification of
  malarial parasite from giemsa-stained thin film images.
\newblock IEEE Access \textbf{8}, 78663--78682 (2020)

\bibitem{laiadi2019kinship}
Laiadi, O., Ouamane, A., Benakcha, A., Taleb-Ahmed, A., Hadid, A.: Kinship
  verification based deep and tensor features through extreme learning machine.
\newblock In: 2019 14th IEEE International Conference on Automatic Face \&
  Gesture Recognition (FG 2019), pp. 1--4. IEEE (2019)

\bibitem{SphereFace}
Liu, W., Wen, Y., Yu, Z., Li, M., Raj, B., Song, L.: Sphereface: Deep
  hypersphere embedding for face recognition.
\newblock In: Proc. of the IEEE Conference on Computer Vision and Pattern
  Recognition, pp. 212--220 (2017)

\bibitem{liu2018conditional}
Liu, Y., Yuan, X., Gong, X., Xie, Z., Fang, F., Luo, Z.: Conditional
  convolution neural network enhanced random forest for facial expression
  recognition.
\newblock Pattern Recognition \textbf{84}, 251--261 (2018)

\bibitem{8599059}
{Lu}, B., {Chen}, J., {Castillo}, C.D., {Chellappa}, R.: An experimental
  evaluation of covariates effects on unconstrained face verification.
\newblock IEEE Transactions on Biometrics, Behavior, and Identity Science
  \textbf{1}(1), 42--55 (2019).
\newblock \doi{10.1109/TBIOM.2018.2890577}

\bibitem{lui2009meta}
Lui, Y.M., Bolme, D., Draper, B.A., Beveridge, J.R., Givens, G., Phillips,
  P.J.: A meta-analysis of face recognition covariates.
\newblock In: 2009 IEEE 3rd International Conference on Biometrics: Theory,
  Applications, and Systems, pp. 1--8. IEEE (2009)

\bibitem{maaten2008visualizing}
Maaten, L.v.d., Hinton, G.: Visualizing data using t-sne.
\newblock Journal of machine learning research \textbf{9}(Nov), 2579--2605
  (2008)

\bibitem{morales21sensitivenets}
Morales, A., Fierrez, J., Vera-Rodriguez, R., Tolosana, R.: {SensitiveNets}:
  Learning agnostic representations with application to face images.
\newblock IEEE Trans. on Pattern Analysis and Machine Intelligence
  \textbf{43}(6), 2158--2164 (2021)

\bibitem{OTOOLE2012169}
O'Toole, A.J., Phillips, P.J., An, X., Dunlop, J.: Demographic effects on
  estimates of automatic face recognition performance.
\newblock Image and Vision Computing \textbf{30}(3), 169 -- 176 (2012).
\newblock \doi{https://doi.org/10.1016/j.imavis.2011.12.007}

\bibitem{pasupa2018virtual}
Pasupa, K., Kudisthalert, W.: Virtual screening by a new clustering-based
  weighted similarity extreme learning machine approach.
\newblock Plos one \textbf{13}(4), e0195478 (2018)

\bibitem{patel2012dictionary}
Patel, V.M., Wu, T., Biswas, S., Phillips, P.J., Chellappa, R.:
  Dictionary-based face recognition under variable lighting and pose.
\newblock IEEE Transactions on Information Forensics and Security
  \textbf{7}(3), 954--965 (2012)

\bibitem{patel20qid}
Perera, P., Fierrez, J., Patel, V.: Quickest intruder detection for multiple
  user active authentication.
\newblock In: IEEE Intl. Conf. on Image Processing (ICIP) (2020).
\newblock \doi{https://arxiv.org/abs/2006.11921}

\bibitem{phillips2011other}
Phillips, P.J., Jiang, F., Narvekar, A., Ayyad, J., O'Toole, A.J.: An
  other-race effect for face recognition algorithms.
\newblock ACM Transactions on Applied Perception (TAP) \textbf{8}(2), 1--11
  (2011)

\bibitem{patel18spm}
Ranjan, R., Sankaranarayanan, S., Bansal, A., Bodla, N., Chen, J.C., Patel,
  V.M., Castillo, C.D., Chellappa, R.: Deep learning for understanding faces:
  Machines may be just as good, or better, than humans.
\newblock IEEE Signal Processing Magazine \textbf{35}(1), 66--83 (2018).
\newblock \doi{10.1109/MSP.2017.2764116}

\bibitem{serna2019algorithmic}
Serna, I., Morales, A., Fierrez, J., Cebrian, M., Obradovich, N., Rahwan, I.:
  Algorithmic discrimination: Formulation and exploration in deep
  learning-based face biometrics.
\newblock arXiv preprint arXiv:1912.01842  (2019)

\bibitem{serna2020sensitiveloss}
Serna, I., Morales, A., Fierrez, J., Cebrian, M., Obradovich, N., Rahwan, I.:
  {SensitiveLoss}: Improving accuracy and fairness of face representations with
  discrimination-aware deep learning.
\newblock arXiv preprint arXiv:2004.11246  (2020)

\bibitem{serna21insidebias}
Serna, I., Peña, A., Morales, A., Fierrez, J.: {InsideBias}: Measuring bias in
  deep networks and application to face gender biometrics.
\newblock In: IAPR Intl. Conf. on Pattern Recognition (ICPR) (2021).
\newblock \doi{https://arxiv.org/abs/2004.06592}

\bibitem{shi2006effective}
Shi, J., Samal, A., Marx, D.: How effective are landmarks and their geometry
  for face recognition.
\newblock Computer Vision and Image Understanding \textbf{102}(2), 117--133
  (2006)

\bibitem{siegel1956nonparametric}
Siegel, S.: Nonparametric statistics for the behavioral sciences.
\newblock McGraw-Hill (1956)

\bibitem{sixta2020fairface}
Sixta, T., Junior, J.C.J., Buch-Cardona, P., Vazquez, E., Escalera, S.:
  Fairface challenge at eccv 2020: analyzing bias in face recognition.
\newblock In: European Conference on Computer Vision, pp. 463--481. Springer
  (2020)

\bibitem{sun2013hybrid}
Sun, Y., Wang, X., Tang, X.: Hybrid deep learning for face verification.
\newblock In: Proceedings of the IEEE International Conference on Computer
  Vision, pp. 1489--1496 (2013)

\bibitem{sun2014deep}
Sun, Y., Wang, X., Tang, X.: Deep learning face representation from predicting
  10,000 classes.
\newblock In: Proceedings of the IEEE Conference on Computer Vision and Pattern
  Recognition, pp. 1891--1898 (2014)

\bibitem{ter21bias}
Terhörst, P., Kolf, J.N., Huber, M., Kirchbuchner, F., Damer, N., Morales, A.,
  Fierrez, J., Kuijper, A.: A comprehensive study on face recognition biases
  beyond demographics.
\newblock arXiv:2103.01592  (2021)

\bibitem{thomee2015new}
Thomee, B., Shamma, D.A., Friedland, G., Elizalde, B., Ni, K., Poland, D.,
  Borth, D., Li, L.J.: The new data and new challenges in multimedia research.
\newblock arXiv preprint arXiv:1503.01817 \textbf{1}(8) (2015)

\bibitem{tome15soft}
Tome, P., Vera-Rodriguez, R., Fierrez, J., Ortega-Garcia, J.: Facial soft
  biometric features for forensic face recognition.
\newblock Forensic Science International \textbf{257}, 171--284 (2015).
\newblock \doi{http://dx.doi.org/10.1016/j.forsciint.2015.09.002}

\bibitem{Vera-Rodriguez_2019_CVPR_Workshops}
Vera-Rodriguez, R., Blazquez, M., Morales, A., Gonzalez-Sosa, E., Neves, J.C.,
  Proenca, H.: Facegenderid: Exploiting gender information in dcnns face
  recognition systems.
\newblock In: Proceedings of the IEEE/CVF Conference on Computer Vision and
  Pattern Recognition (CVPR) Workshops (2019)

\bibitem{vera2019facegenderid}
Vera-Rodriguez, R., Blazquez, M., Morales, A., Gonzalez-Sosa, E., Neves, J.C.,
  Proen{\c{c}}a, H.: Facegenderid: exploiting gender information in dcnns face
  recognition systems.
\newblock In: Proceedings of the IEEE Conference on Computer Vision and Pattern
  Recognition Workshops, pp. 0--0 (2019)

\bibitem{wang2020mitigating}
Wang, M., Deng, W.: Mitigating bias in face recognition using skewness-aware
  reinforcement learning.
\newblock In: Proceedings of the IEEE/CVF Conference on Computer Vision and
  Pattern Recognition, pp. 9322--9331 (2020)

\bibitem{wang2019racial}
Wang, M., Deng, W., Hu, J., Tao, X., Huang, Y.: Racial faces in the wild:
  Reducing racial bias by information maximization adaptation network.
\newblock In: Proceedings of the IEEE International Conference on Computer
  Vision, pp. 692--702 (2019)

\bibitem{wang2019benchmarking}
Wang, Q., Guo, G.: Benchmarking deep learning techniques for face recognition.
\newblock Journal of Visual Communication and Image Representation \textbf{65},
  102663 (2019)

\bibitem{TripletLoss}
Weinberger, K.Q., Saul, L.K.: Distance metric learning for large margin nearest
  neighbor classification.
\newblock Journal of Machine Learning Research \textbf{10}, 207--244 (2009)

\bibitem{wen2016discriminative}
Wen, Y., Zhang, K., Li, Z., Qiao, Y.: A discriminative feature learning
  approach for deep face recognition.
\newblock In: European Conference on Computer Vision, pp. 499--515. Springer
  (2016)

\bibitem{wong2019realization}
Wong, S.Y., Yap, K.S., Zhai, Q., Li, X.: Realization of a hybrid locally
  connected extreme learning machine with deepid for face verification.
\newblock IEEE Access \textbf{7}, 70447--70460 (2019)

\bibitem{yuan2017convolutional}
Yuan, L., Qu, Z., Zhao, Y., Zhang, H., Nian, Q.: A convolutional neural network
  based on tensorflow for face recognition.
\newblock In: 2017 IEEE 2nd Advanced Information Technology, Electronic and
  Automation Control Conference (IAEAC), pp. 525--529. IEEE (2017)

\end{thebibliography}

\end{document}